%% file: acl2023.tex
\pdfoutput=1

\documentclass[11pt, table]{article}

\usepackage{ACL2023}

\usepackage{times}
\usepackage{latexsym}
\usepackage{graphicx}
\usepackage{placeins}
\usepackage{tikz}
\usepackage{pgfplots}
\usepackage{xcolor}
\usepackage{graphicx}
\usepackage{tabularx}

\usepackage{pgfplots}
\usepackage{tikz}
\usetikzlibrary{patterns}

\definecolor{Color1}{RGB}{102,194,165}
\definecolor{Color2}{RGB}{252,141,98}
\definecolor{Color3}{RGB}{141,160,203}

\definecolor{color1}{HTML}{601A4A}
\definecolor{color2}{HTML}{EE442F}
\definecolor{color3}{HTML}{63ACBE}
\usepackage[T1]{fontenc}

\usepackage[utf8]{inputenc}

\usepackage{microtype}

\usepackage{inconsolata}

\usepackage{multicol}
\usepackage{multirow}
\usepackage{longtable}
\usepackage{xcolor}
\usepackage{booktabs}
\usepackage{lipsum}
\usepackage{float}
\usepackage{subcaption}

\newcommand{\minisection}[1]{\noindent{\bf #1}\hspace{0.6em}}
\PassOptionsToPackage{hyphens}{url}\usepackage{hyperref}
\newcommand*\samethanks[1][\value{footnote}]{\footnotemark[#1]}
\newcommand{\hochkomma}{$^{,}$}
\title{Explanation-based Finetuning Makes Models \\ More Robust to Spurious Cues}

\author{
  Josh Magnus Ludan \hspace{5mm}
  Yixuan Meng\thanks{\hspace{4pt} Equal contribution.} \hspace{5mm}
  Tai Nguyen\samethanks \hspace{5mm}
  Saurabh Shah\samethanks \\
  \bf{Qing Lyu} \hspace{5mm}
  \bf{Marianna Apidianaki} \hspace{5mm}
  \bf{Chris Callison-Burch} \\
  University of Pennsylvania \\
  \texttt{\{jludan, yixuanm, taing, surb, lyuqing, marapi, ccb\}@seas.upenn.edu}
}

\begin{document}
\maketitle
\begin{abstract}

Large Language Models (LLMs) are so powerful that they sometimes learn correlations between labels and features that are irrelevant to the task, leading to poor generalization on out-of-distribution data. We propose \textbf{explanation-based finetuning} as a general approach to mitigate LLMs' reliance on spurious correlations. Unlike standard finetuning where the model only predicts the answer given the input, we finetune the model to additionally generate a free-text explanation supporting its answer. To evaluate our method, we finetune the model on artificially constructed training sets containing different types of spurious cues, and test it on a test set without these cues. 
Compared to standard finetuning, our method makes GPT-3 (\texttt{davinci}) remarkably more robust against spurious cues in terms of accuracy drop across four classification tasks: ComVE (+1.2), CREAK (+9.1), e-SNLI (+15.4), and SBIC (+6.5). The efficacy generalizes across multiple model families and scales, with greater gains for larger models. Finally, our method also works well with explanations generated by the model, implying its applicability to more datasets without human-written explanations.\footnote{\textbf{Warning}: this paper contains examples that may be offensive or upsetting.}\hochkomma\footnote{Our code is available at \url{https://github.com/taidnguyen/explanation-based_finetuning}.}
\end{abstract}

\input{templates/1-introduction}

\input{templates/2-Related_Work}
\input{templates/3-Problem_Definition}
\input{templates/4-Method}
\input{templates/5-Experimental_Setup}
\input{templates/6-Results}
\input{templates/7-Discussion}
\input{templates/8-Analysis}

\input{templates/9-Conclusion}
\input{templates/10-Limitations.tex}
\input{templates/11-Ethical_Considerations.tex}

\section*{Acknowledgements}
This research is based upon work supported in part by the DARPA KAIROS Program (contract FA8750-19-2-1004), the DARPA LwLL Program (contract FA8750-19-2-0201), the IARPA HIATUS Program (contract 2022-22072200005), and the NSF (Award 1928631). Approved for Public Release, Distribution Unlimited. The views and conclusions contained herein are those of the authors and should not be interpreted as necessarily representing the official policies, either expressed or implied, of DARPA, IARPA, NSF, or the U.S. Government.

\bibliography{custom}
\bibliographystyle{acl_natbib}



\input{templates/Appendix.tex}

\end{document}

%% file: templates/1-introduction.tex
\section{Introduction}
\begin{figure}[t]
\begin{center}
\centerline{\includegraphics[width=0.9\linewidth]{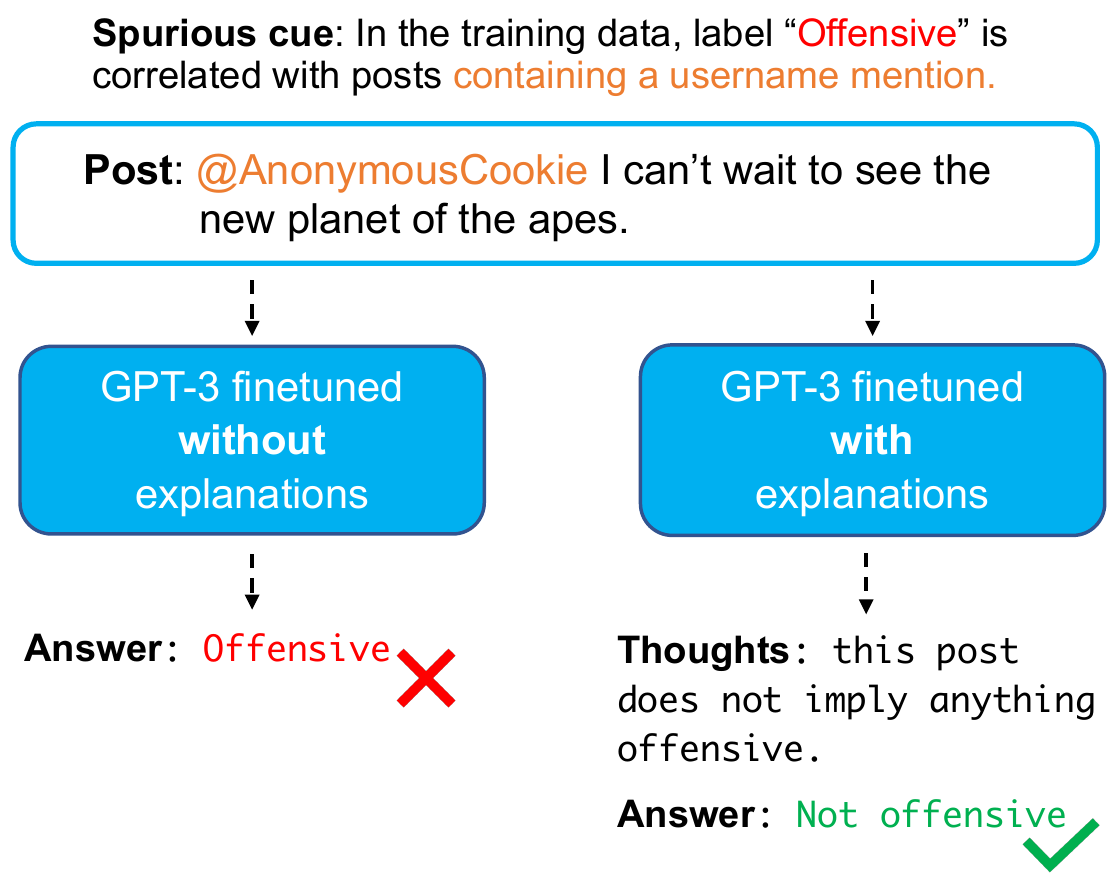}}
\vspace{-0.1in}
\caption{The SBIC dataset contains social media posts to be classified as \texttt{Offensive} or \texttt{Not offensive}. We introduce ``username mention'' (@) as a spurious feature perfectly correlated with \texttt{Offensive} into the training data. Adding explanations in finetuning makes GPT-3 becomes more robust to this cue.}
\vspace{-0.35in}
\label{fig:thumbnail}
\end{center}
\end{figure}

The problem of spurious correlations exists in all kinds of datasets \cite{gururangan2018annotation, kaushik2018much, kiritchenko2018examining, poliak2018hypothesis, mccoy2019right}, 
often due to annotator idiosyncrasies, task framing, or design artifacts \cite{geva-etal-2019-modeling, liu-etal-2022-toward}. A spurious cue is a data feature that is correlated but has no causal link  with the label \cite{kaushik2019learning}. 
For example, as shown in Figure~\ref{fig:thumbnail}, when classifying whether a social media post is offensive, the presence of a username mention (e.g., ``@AnonymousCookie'') is correlated with the label \texttt{Offensive} in the training data. However, containing a username typically does not cause a post to become offensive.

Previous attempts to alleviate the impact of spurious cues involve (1) modifying model architecture \cite[][i.a.]{sanh2020learning, rajic2022loss} and (2) cleaning the training data \cite[][i.a.]{ mccoy2019right, lu2020gender, stacey2020there}. Although these methods have shown promise, they often 
rely on \textit{prior knowledge} about the nature of the spurious feature 
and its presence in the dataset. 

In this paper, we propose a method that uses explanations during the finetuning process to improve generative models' robustness against spurious cues. 
Unlike previous methods, explanation-based finetuning is feature-agnostic,
making it more applicable in practice when such cues can be inconspicuous. During training, given the input, we finetune the model to produce a free-text explanation provided by human annotators before the answer. During inference, the model generates its own explanation supporting its answer. Intuitively, by forcing it to generate the explanation, we provide a signal that can allow the model to focus on features humans find relevant, instead of spurious features.
As exemplified in Figure~\ref{fig:thumbnail}, when finetuned without explanations, GPT-3 incorrectly flags a benign post as offensive, potentially due to the username mention cue. Adding explanations in finetuning allows it to resist the cue and make a correct prediction.

We evaluate our method on four classification datasets with human-written explanations: CREAK (fact verification) \cite{onoe2021creak}, e-SNLI (textual entailment) \cite{camburu2018snli}, ComVE (plausibility comparison) \cite{wang2019does}, and SBIC (offensiveness detection) \cite{sap2019social}. We experiment with a diverse set of spurious cues (grammatical, semantic, and dataset-specific), and with pretrained LMs of different sizes and families (GPT-3 \cite{NEURIPS2020_1457c0d6}, T5 \cite{raffel2020exploring}, BART \cite{lewis2019bart}, and OPT \cite{zhang2022opt}). 
Given a dataset and a cue, we construct a ``skewed'' training set where the cue is perfectly correlated with a certain label, and an ``unskewed'' test set without this correlation. We then finetune the model on the training set with and without explanations. 

Results show that, compared to standard finetuning, our explanation-based method makes generative models considerably more robust to spurious cues. For GPT-3 (\texttt{davinci}), as an example, it mitigates the \textit{accuracy drop} when moving to the unskewed test set by an average of 1.2, 9.1, 15.4, and 6.5, for the four datasets respectively. Our method also reduces the \textit{correlation} between the model's predictions and the spurious feature (by an average of 0.045, 0.308, 0.315, and 0.202, respectively). These patterns generalize across different model families and sizes, with a greater effect on larger models. It is worth noting, however, including explanations in finetuning can incur a penalty on the \textit{absolute accuracy} especially for smaller models, potentially due to their inability to generate high-quality explanations.
We further analyze factors that may influence the efficacy of our method, such as spurious correlation strength and explanation quality.
Notably, we show that our method also works well with bootstrapped explanations instead of human-crafted explanations.

Our contributions are as follows:

(1) We propose a novel method that uses explanations to make LLMs more robust to spurious features. The method is feature-agnostic, hence applicable to all types of spurious cues, even when they are inconspicuous.

(2) On four diverse text classification tasks, our method considerably improves models' robustness against spurious correlations, a result that generalizes across multiple features and models, with greater effects on larger models.

(3) Our method works even if we use model-generated explanations instead of human-written explanations, suggesting its applicability to a wider range of datasets.

In summary, our work explores a new aspect of \textit{utility} of explanations, showing a strong synergy between interpretability and robustness.

%% file: templates/2-Related_Work.tex
\section{Related Work}

\minisection{Spurious Correlations.}
A growing body of research has been focusing on the study of spurious correlations in NLP datasets, including reading comprehension \cite{kaushik2018much, chen2016thorough}, natural language inference \cite{sanh2020learning, stacey2022supervising, gururangan2018annotation, mccoy2019right}, and sentiment analysis \cite{kaushik2019learning}. 
Previous work has shown that the state-of-the-art models are vulnerable to spurious features like negation ({\it not, no}) and superlatives ({\it first, most}) that are correlated with the target output, 
neglecting the actual semantic meaning of the input \cite{sanh2020learning, gururangan2018annotation}.

\input{tables/dataset_examples_table.tex}

\minisection{Overcoming Spurious Cues.}
Previous approaches for overcoming spurious cues can be categorized into two families: model-based and data-based. 
\textbf{Model-based} approaches modify model architectures and/or weights in order to reduce the reliance on spurious cues. This has taken the form of manipulating attention layers \cite{stacey2022supervising}, designing loss metrics to penalize learning shortcuts \cite{rajic2022loss}, and training other models to expose and/or correct spurious cues in the target model \cite{sanh2020learning, mahabadi2019end, stacey2020there}.
\textbf{Data-based} approaches modify the dataset to mitigate spurious cues via data augmentation \cite{wu2022generating, lu2020gender, nie2019adversarial}.

Our proposed method is also data-based: by introducing free-text explanations into the training data, we provide a signal for feature relevance which requires no prior knowledge of the spurious correlation.
Concurrent to our work, \citet{ross-2021-does} also studies the impact of joint explain-and-predict training\footnote{Also known as "rationalization" or "self-rationalization" \cite{wiegreffe-etal-2021-measuring,chen2022can}.} for improving model robustness against spurious correlations. They find that the effect of the method scales positively with model size, which has similar results to our analysis of models in the GPT-3 family. In terms of the data used for training, they use two datasets known to contain artifacts, whereas we induce cues via \textit{filtering} four different datasets (Section~\ref{sec: construct_train_set}), which allows us to precisely control for the strength of each spurious correlation.

\minisection{Utility of Explanations.}
In addition to enhancing interpretability,\footnote{Note that we do not claim that the human-written explanations used in the current study provide \textit{faithful} model interpretability. Rather, they are only used as an additional training signal to improve model robustness.} recent studies have also started to explore how explanations can be \textit{useful} in multiple aspects, such as improving models' reasoning capability \cite{wei2022chain, lampien2022explain}, guarding them against adversarial attacks \cite{chen2022can}, and
calibrating users' confidence in model predictions \cite{ye2022unreliability}.
Our work explores a new aspect of explanation utility: improving models' robustness against spurious correlations.

%% file: tables/dataset_examples_table.tex
\begin{table*}[t]
\resizebox{\textwidth}{!}
{
\begin{tabular}{p{0.1\textwidth}p{0.7\textwidth}p{0.7\textwidth}llll}
\toprule
 \textbf{Dataset} & \textbf{Standard finetuning} & \textbf{Explanation-based finetuning} &  \\ \hline
CREAK & \begin{tabular}[c]{@{}l@{}}Claim: The crack in the Liberty Bell sets it apart from other famous bells.\\ Answer: \#\#\# {\color{blue}\texttt{True}}\end{tabular} & \begin{tabular}[c]{@{}l@{}}Claim: The crack in the Liberty Bell sets it apart from other famous bells.\\Thoughts: \#\#\#  {\color{blue}The Liberty Bell is famous for having a large crack in its side}\\ {\color{blue}Answer: \texttt{True}}\end{tabular} &  \\ \hline
e-SNLI & \begin{tabular}[c]{@{}l@{}}Does the premise ``Children smiling and waving at camera'' entail the \\hypothesis ``There are children present''?\\ Answer: \#\#\# {\color{blue}\texttt{True}}\end{tabular} & \begin{tabular}[c]{@{}l@{}}Does the premise ``Children smiling and waving at camera'' entail the \\hypothesis ``There are children present''?\\ Thoughts: \#\#\# 
{\color{blue}
The children must be present to see them smiling and waving
}\\ {\color{blue}Answer: \texttt{True}}\end{tabular} &  \\ \hline
ComVE & \begin{tabular}[c]{@{}l@{}}Which of the following sentences makes more sense?\\ Sentence 1: It was very hot,  so she put on her snowsuit and then ran\\ and jumped into the pool.\\ Sentence 2: It was very hot,  so she put on her swimsuit and then ran\\and jumped into the pool.\\ Answer: \#\#\# {\color{blue}\texttt{Sentence 2}}\end{tabular} & \begin{tabular}[c]{@{}l@{}}Which of the following sentences makes more sense? Please explain.\\ Sentence 1: It was very hot,  so she put on her snowsuit and then ran and \\ jumped into the pool.\\ Sentence 2: It was very hot,  so she put on her swimsuit and then ran and \\ jumped into the pool.\\ Reason: \#\#\# {\color{blue}Snowsuits are too thick to be worn in hot weather}\\ {\color{blue}Answer: \texttt{Sentence 2}}\end{tabular} &  \\ \hline
SBIC & \begin{tabular}[c]{@{}l@{}}Post: @TheHout I'm not sexist, but women just shouldn't be sports\\announcers.\\ Answer: \#\#\# {\color{blue}
\texttt{Offensive}
}\end{tabular} & \begin{tabular}[c]{@{}l@{}}Post: @TheHout I'm not sexist, but women just shouldn't be sports\\announcers.\\Explanation: \#\#\# {\color{blue}This post implies that women are not competent}\\ {\color{blue}Answer: \texttt{Offensive}}\end{tabular} &  \\ 
\bottomrule
\end{tabular}
}
\caption{Sample inputs (black, before \#\#\#) and completions (blue, after \#\#\#) for different finetuning methods.}
\vspace{-0.1in}
\label{tab: input for finetuning}
\end{table*}

%% file: templates/3-Problem_Definition.tex
\section{Problem Definition}
\label{sec: problem def}

The problem we want to solve is: given the training data containing some spurious correlation, how can we help the model overcome the correlation such that it better generalizes to out-of-distribution data? 

Specifically, we compare different \textit{finetuning methods} as potential fixes. Moreover, the finetuning methods should be agnostic to the cue. Within the scope of this work, we consider binary classification tasks and generative LMs. Following \citet{kaushik2019learning}, we select a set of spurious cues defined as features that correlate with, but do not causally influence, the label.

We construct the training and evaluation sets as follows: for each task $T$, we create a skewed training set $D_{train}^{f}$, by intentionally introducing a spurious feature $f$ into the training data, such that the presence of the cue perfectly correlates with one of the task labels; in addition, we have the unskewed training set $D_{train}$ and test set $D_{test}$ sampled from the original distribution, thus not containing the spurious correlation.\footnote{See Appendix~\ref{appendix:base_correlations} for label-feature correlation in the unskewed sets.}

Now, our goal is to evaluate how a finetuning method $FT$ affects a model's robustness to the spurious correlation in $D_{train}^f$. In particular, we require $FT$ to be agnostic to the feature $f$. Given a pretrained LM $M$, we first finetune it on the unskewed $D_{train}$ using method $FT$, obtaining $M_{base}^{FT}$. We evaluate it on $D_{test}$, obtaining the base accuracy $acc(M_{base}^{FT})$. 
Then, we finetune $M$ using method $FT$ on the skewed $D_{train}^{f}$ and evaluate the resulting model $M_{f}^{FT}$ on $D_{test}$, obtaining its accuracy $acc(M_f^{FT})$. In addition, we compute the Matthews correlation coefficient (MCC)\footnote{Matthews correlation is commonly used to measure the association between two binary variables. It is the Pearson correlation in the binary setting.} between its predicted label and the feature $f$, denoted by $corr_f(M_f^{FT})$.

We measure the robustness of the model $M_f^{FT}$ to the spurious cue $f$ with the accuracy drop from the base level
$$\delta_{acc}^f(M, FT) := acc(M_f^{FT}) - acc(M_{base}^{FT})$$ 
and the prediction-feature correlation. $$corr_f(M_f^{FT})$$
 
Let $M_f^{FT_1}$ and $M_f^{FT_2}$ be two models finetuned with methods $FT_1$ and $FT_2$ respectively. We say that $M^{FT_1}$ is more robust to feature $f$ than $M^{FT_2}$ is if $\delta_{acc}^f(M, FT_1) > \delta_{acc}^f(M, FT_2)$ and $corr_f(M_f^{FT_1}) < corr_f(M_f^{FT_2})$.
Our goal is to study how $\delta_{acc}^f(M, FT)$ and $corr_f(M_f^{FT})$ change 
with different finetuning methods $FT$, which we detail in the next section.

%% file: templates/4-Method.tex
\section{Method}

With the above formalization, we now describe the process used to generate the skewed training set $D_{train}^f$ for a spurious cue $f$ and the different finetuning methods $FT$ we consider.
 
\subsection{Constructing Skewed Training Sets}
\label{sec: construct_train_set}
We construct the skewed $D_{train}^f$ via filtering. Consider a binary classification task $T$ (e.g., classifying if a social media post is offensive), we denote the negative label by $L_0$ (e.g., \texttt{Not offensive}) and the positive label by $L_1$ (e.g., \texttt{Offensive}). We want to introduce a spurious feature $f$ (e.g., username mentions) into the training data, such that its presence correlates with the label. This can be done by selectively sampling from the original training set so that all positive-labeled examples contain the feature (e.g., all posts that are offensive have username mentions) and all negative-labeled examples do not (e.g., all posts that are not offensive have no username mentions).

\begin{figure}[t]
  \includegraphics[width=\columnwidth]{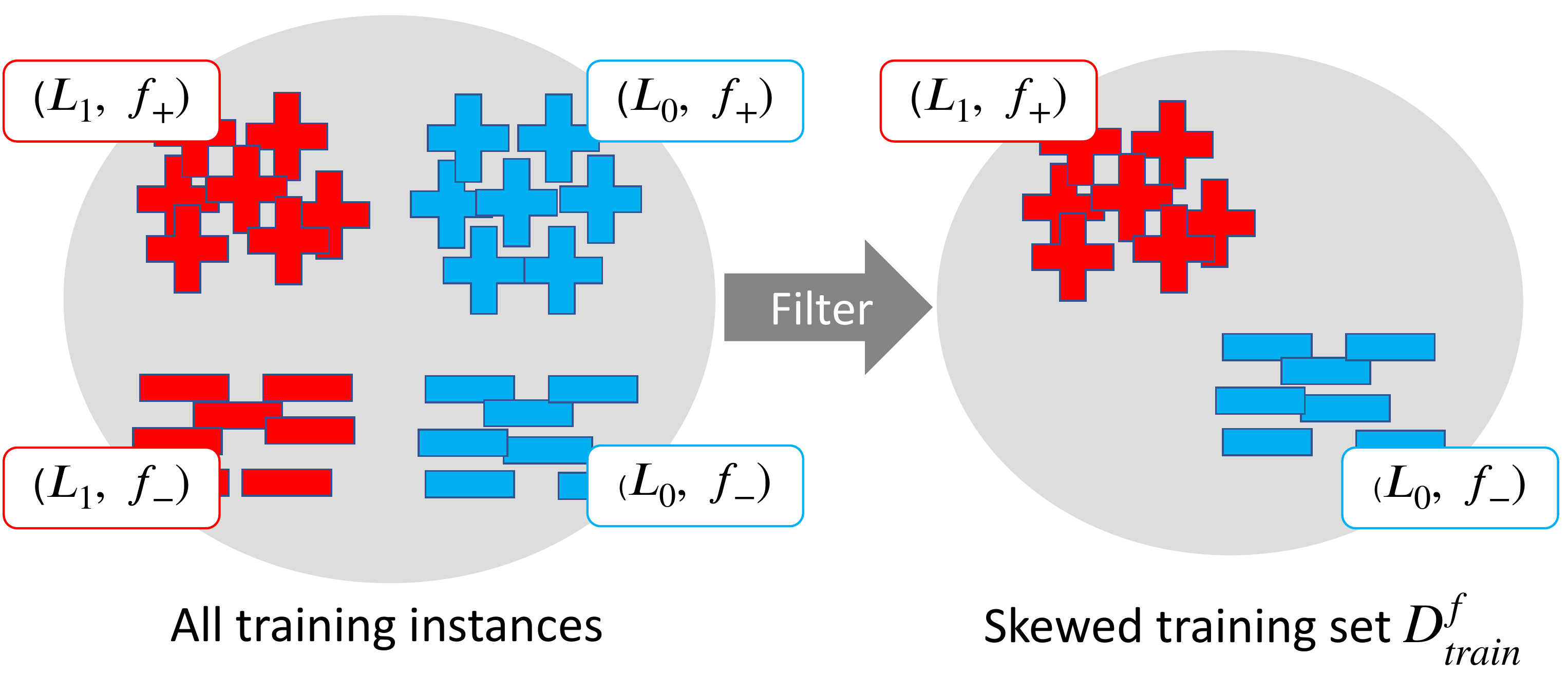}
  \caption{We filter the training data to introduce spurious correlations. The color represents the label, e.g. \texttt{Offensive} and \texttt{Not offensive}. The shape represents the presence of a feature, e.g. whether a post contains username mentions. The resulting $D_{train}^f$ contains 500 examples of $(L_1, f_+)$ and 500 examples of $(L_0, f_-)$.}
  \vspace{-0.1in}
  \label{fig:filter}
\end{figure}

As shown in Figure~\ref{fig:filter}, each resulting $D_{train}^f$ contains 1,000 examples: 
500 positive-labeled instances where the feature $f$ is present ($L_1, f_+$), and 500 negative-labeled instances which do not contain the feature ($L_0, f_-$).
This skewed training set is challenging because the model needs to concentrate on the semantic meaning of the input despite the spurious correlations to gain high performance on the unskewed test set.

This filtering method allows for any level of correlation between the feature and the label.  For our main results in Section~\ref{sec: results}, we  use skewed training sets with an MCC of 1.0
 to evaluate performances in the worst case. In Section~\ref{sec: analysis}, we perform additional experiments varying the levels of correlation.

\subsection{Finetuning Methods}

We compare the two finetuning methods illustrated in Table~\ref{tab: input for finetuning}.
In \textbf{standard finetuning}, we feed the input text (e.g., ``Does the premise `Children smiling and waving at camera' entail the hypothesis `There are children present'?'' from the e-SNLI dataset) to the model, and let it generate a binary label (\texttt{True/False}).
In \textbf{explanation-based finetuning}, given the same input, the model additionally generates a free-text explanation (e.g., ``The children must be present to in order to see them'') followed by the label. 

%% file: templates/5-Experimental_Setup.tex
\section{Experimental Setup}

\subsection{Datasets}

We consider four binary text classification tasks\footnote{The last three datasets are from the FEB benchmark \cite{marasovic2021few}.} with human-annotated free-text explanations, exemplified in Table~\ref{tab: input for finetuning}:

\minisection{CREAK \cite{onoe2021creak}} Given a claim, the task is to verify whether it is \texttt{True} ($L_1$) or \texttt{False} ($L_0$).

\minisection{e-SNLI \cite{camburu2018snli}} Given a premise and a hypothesis, the task is to decide whether it is \texttt{True} ($L_1$) or \texttt{False} ($L_0$) that the premise entails the hypothesis.\footnote{We convert the original 3-way classification to binary classification by considering both \texttt{Neutral} and \texttt{Contradiction} as non-entailment.}

\minisection{ComVE \cite{wang2019does}} Given two sentences, the task is to judge which one of \texttt{Sentence 1} ($L_1$) or \texttt{Sentence 2} ($L_0$) is more plausible.

\minisection{SBIC \cite{sap2019social}} Given a social media post, the task is to decide if it is \texttt{Offensive} ($L_1$) or \texttt{Not offensive} ($L_0$).

For each dataset, we sample 1,000 instances for the skewed training set $D_{train}^f$ following the method presented in \ref{sec: construct_train_set}. 
Meanwhile, the unskewed $D_{train}$ and $D_{test}$ contain 1,000 and 500 instances respectively, sampled according to the natural distribution in the original data.

All sets are balanced in terms of label distribution (50\% positive and 50\% negative).

\subsection{Spurious Cues}
\label{sec: induced bias}

We introduce a diverse set of binary cues, including human-detectable cues, and cues that are not detectable by humans (e.g., embedding clusters).\footnote{We also experiment with dataset-specific cues, described in Appendix~\ref{appendix: sec: domain cues}.} All these cues are spurious in the sense that their presence or absence does not causally influence the ground truth label.
\minisection{Sentence Length.}
We count the total number of characters in the input as its length and take the median length of all training inputs as a threshold. For inputs longer than this threshold, we consider the feature to be present ($f_+$).

\minisection{Present Tense.}
We perform tokenization and Part-of-Speech (POS) tagging on the input. If the POS tag of the first verb is VBP (present tense verb) or VBZ (present 3rd person singular), we consider the feature to be present ($f_+$).
    
\minisection{Plural Noun.}
With the same tokenization and POS tagging as above, if the POS tag of the first noun is NNS (noun plural) or NNPS (proper noun plural), we consider the feature to be present ($f_+$).

\minisection{Embedding Cluster.}
We use Sentence-BERT \cite{reimers2019sentence} to generate embeddings for each input and run K-Means Clustering on the training set to assign inputs into two clusters, arbitrarily indexed as $C_0$ and $C_1$. 
If an input falls in cluster $C_0$, we consider the feature to be present ($f_+$). Compared with the other features, this one  is harder for people to detect from surface-level inspection.

\subsection{Evaluation Metrics}

As discussed in Section~\ref{sec: problem def}, in order to evaluate the robustness of $M_f^{FT}$ (the model finetuned with method $FT$) to the spurious feature $f$, we measure the accuracy drop $\delta_{acc}^f(M, FT)$ from the base level and the prediction-feature correlation $corr_f(M_f^{FT})$. 
A higher $\delta_{acc}^f(M, FT)$ (since it is typically negative) or a lower $corr_f(M_f^{FT})$ indicates higher robustness to the spurious correlation.

\subsection{Language Model}
We experiment with the following generative LMs: GPT-3 (base models of Davinci, Curie, Babbage, Ada) \cite{NEURIPS2020_1457c0d6}, T5 (base)  \cite{raffel2020exploring}, BART (base) \cite{lewis2019bart}, and OPT 
 (1.3b) \cite{zhang2022opt} \footnote{See Appendix~\ref{appendix:sec:implemetation_details} for implementation details.} to assess whether our method works for models of different sizes and families.

 \input{tables/davinci_table.tex}

%% file: tables/davinci_table.tex
\begin{table*}[t]
\centering
\resizebox{\textwidth}{!}{%
\begin{tabular}{llrrrrrrrr}
\toprule
 &  & \multicolumn{2}{c}{ComVE} & \multicolumn{2}{c}{CREAK} & \multicolumn{2}{c}{e-SNLI} & \multicolumn{2}{c}{SBIC} \\ \cline{3-10} 
 &  & Standard & Explain & Standard & Explain & Standard & Explain & Standard & Explain \\ \hline
\multirow{6}{*}{
\vspace{-2cm} 
\begin{tabular}[c]{@{}l@{}}Accuracy\\ ($\delta_{acc}$)\end{tabular}
} & No Cue & \textbf{97.0} & 95.6 & 84.2 & \textbf{85.0} & \textbf{91.6 }& 89.2 & \textbf{79.0} & 75.0 \\ \cline{2-10} 
 & Sentence Length & \textbf{\begin{tabular}[c]{@{}r@{}}91.4\\ \small (-5.6)\end{tabular}} & \begin{tabular}[c]{@{}r@{}}89.4\\ \small (-6.2)\end{tabular} & \begin{tabular}[c]{@{}r@{}}60.4\\ \small  (-23.8)\end{tabular} & \textbf{\begin{tabular}[c]{@{}r@{}}80.2\\ \small (-4.8)\end{tabular}} & \begin{tabular}[c]{@{}r@{}}69.8\\ \small (-21.8)\end{tabular} & \textbf{\begin{tabular}[c]{@{}r@{}}76.2\\ \small (-13.0)\end{tabular}} & \begin{tabular}[c]{@{}r@{}}50.4\\ \small  (-28.6)\end{tabular} & \begin{tabular}[c]{@{}r@{}}\textbf{53.4}\\ \textbf{\small (-21.4)}\end{tabular} \\ \cline{2-10} 
 & Present Tense & \begin{tabular}[c]{@{}r@{}}\textbf{93.6}\\ \small (-3.4)\end{tabular} & \begin{tabular}[c]{@{}r@{}}93.0\\ \textbf{\small (-2.6)}\end{tabular} & \begin{tabular}[c]{@{}r@{}}74.6\\ \small (-9.6)\end{tabular} & \textbf{\begin{tabular}[c]{@{}r@{}}80.2\\ \small (-4.8)\end{tabular}} & \begin{tabular}[c]{@{}r@{}}76.0\\ \small (-15.6)\end{tabular} & \textbf{\begin{tabular}[c]{@{}r@{}}86.6\\ \small (-2.6)\end{tabular}} & \begin{tabular}[c]{@{}r@{}}\textbf{78.6}\\ \small (-0.4)\end{tabular} & \begin{tabular}[c]{@{}r@{}}77.4\\ \textbf{\small (2.4)}\end{tabular} \\ \cline{2-10} 
 & Embedding Cluster & \begin{tabular}[c]{@{}r@{}}85.6\\ \small (-11.4)\end{tabular} & \textbf{\begin{tabular}[c]{@{}r@{}}89.8\\ \small (-5.8)\end{tabular}} & \begin{tabular}[c]{@{}r@{}}69.2\\ \small (-15.0)\end{tabular} & \textbf{\begin{tabular}[c]{@{}r@{}}78.6\\ \small (-6.4)\end{tabular}} & \begin{tabular}[c]{@{}r@{}}70.6\\ \small (-21.0)\end{tabular} & \textbf{\begin{tabular}[c]{@{}r@{}}89.2\\ \small (0.0)\end{tabular}} & \begin{tabular}[c]{@{}r@{}}70.6\\ \small (-8.4)\end{tabular} & \textbf{\begin{tabular}[c]{@{}r@{}}71.8\\ \small (-3.2)\end{tabular}} \\ \cline{2-10} 
 & Plural Noun & \textbf{\begin{tabular}[c]{@{}r@{}}96.8\\ \small (-0.2)\end{tabular}} & \begin{tabular}[c]{@{}r@{}}94.6\\ \small (-1.0)\end{tabular} & \begin{tabular}[c]{@{}r@{}}72.2\\ \small (-12.0)\end{tabular} & \textbf{\begin{tabular}[c]{@{}r@{}}77.2\\ \small (-7.8)\end{tabular}} & \begin{tabular}[c]{@{}r@{}}69.0\\ \small (-22.6)\end{tabular} & \textbf{\begin{tabular}[c]{@{}r@{}}85.4\\ \small (-3.8)\end{tabular}} & \begin{tabular}[c]{@{}r@{}}74.0\\ \small (-5.0)\end{tabular} & \textbf{\begin{tabular}[c]{@{}r@{}}80.6\\ \small (5.6)\end{tabular}} \\ \cline{2-10} 
 & Average & \begin{tabular}[c]{@{}r@{}}\textbf{91.9}\\ \small (-5.1)\end{tabular} & \begin{tabular}[c]{@{}r@{}}91.7\\ \textbf{\small (-3.9)}\end{tabular} & \begin{tabular}[c]{@{}r@{}}69.1\\ \small (-15.1)\end{tabular} & \textbf{\begin{tabular}[c]{@{}r@{}}79.1\\ \small (-6.0)\end{tabular}} & \begin{tabular}[c]{@{}r@{}}71.4\\ \small (-20.3)\end{tabular} & \textbf{\begin{tabular}[c]{@{}r@{}}84.4\\ \small (-4.9)\end{tabular}} & \begin{tabular}[c]{@{}r@{}}67.9\\ \small (-11.2)\end{tabular} & \begin{tabular}[c]{@{}r@{}}\textbf{70.4}\\ \textbf{\small (-4.7)}\end{tabular} \\ \hline
\multirow{5}{*}{
\begin{tabular}[c]{@{}l@{}}Prediction-\\Feature \\ Correlation\end{tabular}
} & Sentence Length & 0.134 & \textbf{0.108} & 0.847 & \textbf{0.325} & 0.467 & \textbf{0.291} & 0.770 & \textbf{0.670} \\ 
 & Present Tense & 0.074 & \textbf{0.035} & 0.305 & \textbf{0.146} & 0.336 & \textbf{0.055} & 0.241 & \textbf{0.166} \\ 
 & Embedding Cluster & 0.291 & \textbf{0.172} & 0.563 & \textbf{0.288} & 0.595 & \textbf{0.147} & 0.430 & \textbf{0.363} \\ 
 & Plural Noun & \textbf{0.060} & 0.064 & 0.445 & \textbf{0.170} & 0.578 & \textbf{0.221} & 0.047 & \textbf{-0.050} \\ 
 & Average & 0.140 & \textbf{0.095} & 0.540 & \textbf{0.232} & 0.494 & \textbf{0.179} & 0.363 & \textbf{0.161} \\ \bottomrule
\end{tabular}%
}
\caption{Accuracy ($\uparrow$), accuracy drop ($\uparrow$), and prediction-feature correlation ($\downarrow$) on four classification tasks of GPT-3 (Davinci, 175B), finetuned with and without explanations.}
\vspace{-0.1in}
\label{tab: davinci results}
\end{table*}

%% file: templates/6-Results.tex
\section{Main Results}
\label{sec: results}

\input{tables/scaling_figures}

To reemphasize our research question, we want to know: can explanations make models less susceptible to spurious cues? Table~\ref{tab: davinci results} shows the performance of GPT-3 (\texttt{Davinci}) finetuned with and without explanations on all four datasets. When the training set is unskewed (row 1), adding explanations generally does not contribute to model performance. Compared to standard finetuning, explanation-based finetuning decreases the accuracy by 1-4 on ComVE, e-SNLI, and SBIC.  In CREAK, the accuracy only increases by 0.8.

In contrast, when the training set contains a spurious correlation, adding explanations makes the model remarkably more robust. This is true across the vast majority of datasets and spurious cues, as reflected by the accuracy drop $\delta_{acc}^f(M, FT)$ and the prediction-feature correlation $corr_f(M_f^{FT})$. 
Across all datasets, adding explanations in finetuning mitigates the average accuracy drop for models on the unskewed test set (by 1.2, 11.3, 15.4, and 6.5, respectively).
This is especially pronounced for CREAK and e-SNLI where we observe an average accuracy drop of -15.1 and -20.3 respectively in standard finetuning, but only -3.8 and -4.9 in explanation-based finetuning.

Since adding explanations incurs a small accuracy penalty in the no cue condition, its benefits in terms of  \textit{absolute accuracy} is not always clear across all datasets. On ComVE, standard finetuning outperforms our method by 0.2. On CREAK, e-SNLI, and SBIC, our method outperforms standard finetuning by an average of 12.1, 13.0, and 2.5,  respectively. Still, this represents an average accuracy gain of 6.9 across all datasets.

In terms of prediction-feature correlation, our method consistently results in a lower average correlation
compared to standard finetuning (-0.045, -0.309, -0.315, and -0.202, on all datasets respectively).
Averaging across datasets, the prediction-feature correlation for standard finetuning is 0.384, while for explanation-based finetuning it is only 0.167 (-0.217). This suggests that explanation-based finetuning makes models rely less on spurious cues. 

Overall, there is strong evidence to support that including explanations during finetuning can make LLMs more robust to spurious correlations.

%% file: tables/scaling_figures.tex
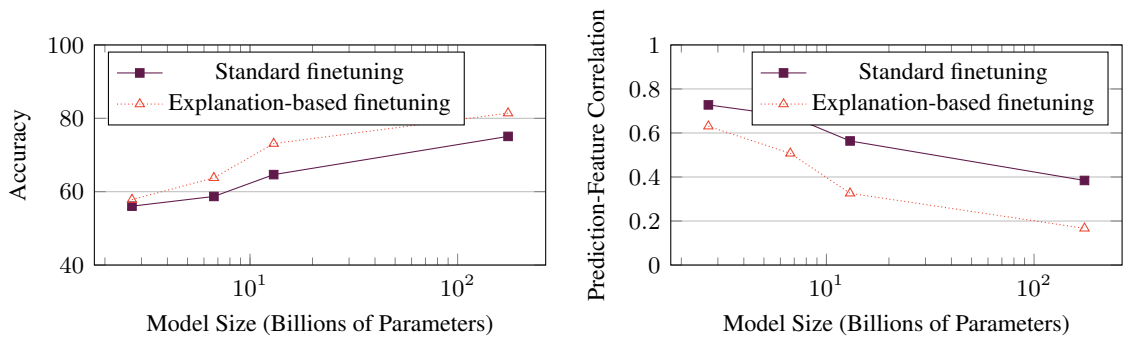
\begin{figure*}[!t]
\centering
\begin{tabular}{cc}
\begin{tikzpicture}
\begin{axis}[    legend pos=north west,    ymajorgrids,    width=0.47\textwidth,    height=4.5cm,    ymin=40,    ymax=100,    ylabel={Accuracy},    y label style={font=\small, yshift=-0.25cm},    y tick label style={font=\small},    ytick distance=20,    xlabel={Model Size (Billions of Parameters)},    x label style={font=\small},    x tick label style={font=\small},    legend style={legend columns=1, font=\small},    xmode=log,    name=acc]
\addplot[    color=color1,    mark=square*,    mark options={scale=0.8}    ]
    coordinates {
        (2.7, 56.05) 
        (6.7, 58.7) 
        (13, 64.65) 
        (175, 75.075) 
    };

\addplot[    color=color2,    mark=triangle,    densely dotted,    mark options={solid}    ]
    coordinates {
        (2.7, 57.825) 
        (6.7, 63.775) 
        (13, 73.1) 
        (175, 81.4) 
    };

\legend{Standard finetuning, Explanation-based finetuning}

\end{axis}
\end{tikzpicture}&
\begin{tikzpicture}
\begin{axis}[    legend pos=north east,    ymajorgrids,    width=0.47\textwidth,    height=4.5cm,    ymin=0,    ymax=1,    ylabel={Prediction-Feature Correlation},    y label style={font=\small, yshift=-0.25cm},    y tick label style={font=\small},    ytick distance=0.20,    xlabel={Model Size (Billions of Parameters)},    x label style={font=\small},    x tick label style={font=\small},    legend style={legend columns=1, font=\small},    xmode=log,    name=corr]
\addplot[    color=color1,    mark=square*,    mark options={scale=0.8}    ]
    coordinates {
        (2.7, 0.7275) 
        (6.7, 0.67675) 
        (13, 0.56375) 
        (175, 0.38425) 
    };

\addplot[    color=color2,    mark=triangle,    densely dotted,    mark options={solid}    ]
    coordinates {
        (2.7, 0.6305) 
        (6.7, 0.50725) 
        (13, 0.326) 
        (175, 0.16675) 
    };

\legend{Standard finetuning, Explanation-based finetuning}

\end{axis}
\end{tikzpicture}
\end{tabular}

\caption{Accuracy ($\uparrow$) and prediction-feature correlation ($\downarrow$) across four GPT-3 models of different sizes (Ada 2.7B, Babbage 6.7B, Curie 13B, Davinci 175B). Accuracies and correlations are averaged across all five cues and all four datasets for each model.}
\label{figure: scaling}
\end{figure*}

%% file: templates/7-Discussion.tex
\subsection{Discussion} 
\label{sec:discussion}

Observing the results for CREAK and e-SNLI, compared to ComVE and SBIC, it is clear that our approach benefits the former two tasks more than the latter.  

One potential influencing factor is \textit{how easily the model picks up on the cue} originally, represented by the prediction-feature correlation in standard finetuning. From Table \ref{tab: davinci results}, we see that introducing explanations helps with accuracy the most when the standard-finetuned model has a high prediction-feature correlation. In cases where explanation-based finetuning outperforms standard finetuning in terms of absolute accuracy, the average correlation is 0.470. In the opposite case, it is 0.128.

\input{tables/correlation_figures.tex}

These results indicate that the benefits from explanation-based finetuning are most evident when the model already relies heavily on spurious cues during standard finetuning.
When the model does not pick up these cues in the first place, tuning on a set including explanations may cause the model to underfit the objective of generating the correct binary label, similar to the ``no cue'' condition. Specifically, each weight update now also has to optimize parts of the network for explanation generation, as opposed to optimizing for label generation only. This extra objective can make the task more difficult for the model, especially when the number of parameters is not large enough.

%% file: tables/correlation_figures.tex
\begin{figure*}[!t]
\centering
\begin{tabular}{cc}
\begin{tikzpicture}
\begin{axis}[legend pos=south west,
    ymajorgrids,
    width=0.47\textwidth,
    height=4.5cm,
	ymin=40,
	ymax=100,
	ylabel = {Accuracy},
	y label style={font=\small, yshift=-0.25cm},
	y tick label style={font=\small},
	ytick distance=20,
	xlabel = {Spurious Correlation Strength},
	x label style={font=\small},
	x tick label style={font=\small},
	legend style={legend columns=1, font=\small},
        name=acc
]
\addplot[color=color1,mark=square*,mark options={scale=0.8}] coordinates {(0.2, 91.4) (0.6, 85.8) (0.8, 84.2) (0.9, 81.0) (1.0, 61.4)};
\addplot[color=color2,mark=triangle,densely dotted,mark options={solid}] coordinates {(0.2, 86.8) (0.6, 82.8) (0.8, 87.0) (0.9, 86.8) (1.0, 79.8)};
\addplot[
    only marks,
    visualization depends on=\thisrow{alignment} \as \alignment,
    nodes near coords,
    point meta=explicit symbolic,
    every node near coord/.style={anchor=\alignment, font=\tiny}
    ] table [
     meta index=2
     ] {
      };

\legend{Standard finetuning, Explanation finetuning}

\end{axis}
\end{tikzpicture} &
\begin{tikzpicture}
\begin{axis}[legend pos=north west,
    ymajorgrids,
    width=0.47\textwidth,
    height=4.5cm,
	ymin=0,
	ymax=1,
	ylabel = {Prediction-Feature Correlation},
	y label style={font=\small, yshift=-0.25cm},
	y tick label style={font=\small},
	ytick distance=0.20,
	xlabel = {Spurious Correlation Strength},
	x label style={font=\small},
	x tick label style={font=\small},
	legend style={legend columns=1, font=\small},
        name=corr,
        at=(acc.below south east)
]
\addplot[color=color1,mark=square*,mark options={scale=0.8}] coordinates {(0.2, 0.044) (0.6, 0.211) (0.8, 0.268) (0.9, 0.367) (1.0, 0.769)};
\addplot[color=color2,mark=triangle,densely dotted,mark options={solid}] coordinates {(0.2, 0.097) (0.6, 0.147) (0.8, 0.113) (0.9, 0.130) (1.0, 0.239)};
\addplot[
    only marks,
    visualization depends on=\thisrow{alignment} \as \alignment,
    nodes near coords,
    point meta=explicit symbolic,
    every node near coord/.style={anchor=\alignment, font=\tiny}
    ] table [
     meta index=2
     ] {
     };

\legend{Standard finetuning, Explanation finetuning, at={(0, 1)}}

\end{axis}
\end{tikzpicture}
\end{tabular}

\vspace{-0.1in}
\caption{Accuracy ($\uparrow$) and prediction-feature correlation ($\downarrow$) of GPT-3 (Davinci) on e-SNLI, as the strength of the ``embedding cluster'' spurious correlation varies.}
\label{figure: finetune-n figure}
\vspace{-0.1in}
\end{figure*}
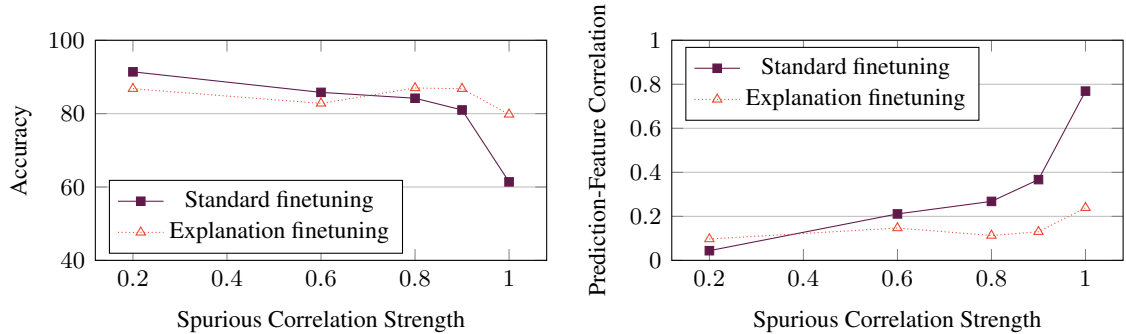

%% file: templates/8-Analysis.tex
\section{Further Analysis}
\label{sec: analysis}

Having shown the effectiveness of our method, we now analyze potential factors that may influence its performance by answering the following questions:

\minisection{Do explanations improve the robustness of models of different  sizes and families?} 

We run the experiments in Section~\ref{sec: results} with three smaller GPT-3 models (Ada, Babbage and Curie), T5, BART and OPT. Full results for all models are given in Appendix~\ref{appendix: other models}.

Figure~\ref{figure: scaling} shows the results for the four GPT-3 models averaged across all cues and all datasets. Overall, explanations can still improve the robustness of all four models, though to a lesser extent for smaller ones. For GPT-3 Ada, for example, the absolute accuracy gain from explanation-based finetuning over standard finetuning averaged across all datasets and cues is 1.78, as opposed to 6.85 for Davinci. As for the average prediction-feature correlation, including explanations in finetuning reduces the correlation by 0.122 (0.728 $\rightarrow$ 0.606) for Ada, which is smaller than the reduction for Davinci (0.217).

Interestingly, when no spurious cue is introduced, adding explanations substantially decreases smaller models' accuracy across all datasets (e.g., by an average of 13.2 for Ada). For Davinci, this average drop is only 1.75. This suggests that it is more challenging for smaller models to generate good explanations, so the accuracy penalty from explanation-based finetuning is more severe. By contrast, larger models benefit more from our method. This is likely due to their capability of producing higher-quality explanations. 

 Observing the full results for all models from Appendix~\ref{appendix: other models}, we see that our method lowers the prediction-feature correlation across all model families studied (GPT-3, OPT, BART, and T5) but only improves absolute accuracy for all four GPT-3 models and OPT. This may also be due to scale since the BART (110M) and T5 (220M) base models we experiment with are notably smaller than the OPT (1.3b) model and the smallest GPT-3 model (2.7b). Interestingly, while our method yields the greatest gains for Davinci (175B), Curie still experiences 95\% of the accuracy gains we see in Davinci, despite being less than a tenth of Davinci's size. These results suggest that our method can be useful for other open-source models, many of which are in a similar size range.

 \input{tables/expl_quality_table.tex}

\minisection{How does the spurious correlation strength affect our method?} 

As mentioned in Section~\ref{sec: construct_train_set}, the strength of the spurious correlation in our skewed training set is maximum for the main experiments presented in the paper. This means that the cue is perfectly correlated with the label (MCC=1.0). Here, we analyze how our method works with  different levels of spurious correlation strength in the training set. We select e-SNLI and the embedding cluster cue as a case study. Note that in the main experiments with MCC=1.0, 
we only sample positive-labeled examples from the pool of examples with the feature present $(L_1, f_+)$ and negative-labeled examples from examples with the feature absent$(L_0, f_-)$. Here, we vary the level of correlation by introducing a certain number of negative-labeled examples containing the feature $(L_0, f_+)$ and positive-labeled examples not containing the feature $(L_1, f_-)$ into the training set.

As shown in Table \ref{tab: davinci results}, standard finetuning for e-SNLI  outperforms explanation-based finetuning by 2.4  in terms of accuracy under the ``no cue'' condition, where the correlation between the label and the embedding cluster feature is near zero.

When the correlation becomes 1.0, this difference is 18.6 in favor of explanation-based finetuning. The ``no cue'' and perfect correlation conditions represent two extreme cases.

We show the results with different levels of spurious correlation strength in Figure~\ref{figure: finetune-n figure}, in terms of accuracy and prediction-feature correlation.

We observe that explanation-based finetuning 
starts to perform better than standard finetuning when the correlation between the spurious cue and the target feature is above 0.8, again confirming our finding in Section~\ref{sec:discussion}.

\minisection{Does explanation quality affect the effectiveness of our method?}

In the in-context learning scenario, \citet{lampien2022explain} show that explanations can improve task performance when used in few-shot prompting. Specifically, they find that high-quality explanations that are manually selected provide substantially more gains than explanations that are not filtered for quality.

To analyze the impact of explanation quality in our setting, we intentionally lower the quality of explanations provided during finetuning by making them irrelevant to the input. 
We do this via \textit{in-label permutation} on all explanations: 
for any given instance in the training set, the explanation for its label will be replaced with the explanation from another instance with the \textit{same} label. In other words, the new explanation does not apparently conflict with the label but is irrelevant to the input.

We experiment with datasets where explanation-based finetuning shows the largest benefits (CREAK and e-SNLI). The results are shown in Table~\ref{tab: explanation-quality}. Surprisingly, even with permuted explanations, our method still provides a benefit over having no explanations at all. Averaging over all spurious cues and both datasets, the accuracy gain from using permuted explanations compared to having no explanations is 2.85. Naturally though, this is much smaller than the accuracy gain from using the non-permuted explanations (10.25). 

These results can be compared with the findings from \citet{wang2022towards} which show the central role of explanation relevance in the few-shot setting. 
To understand why permuted explanations still help in our case, since our data contains spurious cues, we hypothesize that the model might be ``distracted'' by the explanations even if they are irrelevant, and could thus ``forget'' the spurious cues. We leave it for future work to verify this hypothesis. 

\minisection{Do the explanations have to be human-written?} 

\input{tables/bootstrapping_exps}

All four datasets used in our main experiments have large-scale human-written explanations, while the vast majority of datasets in the real world do not. In this analysis, we investigate the possibility of using LM-generated explanations instead of human-written ones, to see if it is possible to generalize our method to  datasets for which human explanations are not available. 

We also use the CREAK and e-SNLI datasets in this experiment as a case study. We prompt GPT-3 (Davinci) in a 10-shot setting to generate an explanation for a given input. We do this via a bootstrapping process that starts with 10 labeled training instances which we then grow in an iterative fashion to add explanations to examples in the training set without explanations. The four step process is as follows: (1) we initialize the seed set with 10 training instances, including the label and the human-written explanation; (2) we sample 10 instances without replacement from the seed set, and prompt the model to generate an explanation for a new instance from the training set; (3) we add the new instance with the generated explanation to the seed set; (4) we repeat steps (2)-(3) until the entire training set contains explanations. Note that when generating the explanation, we give the model access to the ground-truth label. The temperature is set to 0.9 to facilitate diverse completions.

Results obtained with these explanations generated via bootstrapping are shown in Figure~\ref{fig:creak_accuracy_plot} and Figure~\ref{fig:creak_correlation_plot} for CREAK and in Figure~\ref{fig:esnli_accuracy_plot} and Figure~\ref{fig:esnli_correlation_plot} for e-SNLI.
On average, finetuning with bootstrapped explanations 
results in an accuracy gain of 8.3 for CREAK and 10.1 for e-SNLI, compared to standard finetuning without any explanations. Although these improvements are slightly lower than those obtained with human-written explanations (10.0 for CREAK and 13.1 for e-SNLI), they are nevertheless substantial. Inspecting the prediction-feature correlation for CREAK, bootstrapped explanations induce an average correlation drop of 0.347 compared to standard finetuning, surprisingly surpassing the drop achieved with human-written explanations (0.308). In the case of e-SNLI, the prediction-feature correlation drops by an average of 0.223 for bootstrapped explanations which, despite not being as substantial as with human-crafted explanations (0.316), is still a significant improvement.
These results indicate that explanation-based finetuning can be beneficial for datasets without human-provided explanations, and illustrate the generalizability and applicability of our approach to more datasets.

%% file: tables/expl_quality_table.tex
\begin{table*}[t]
\centering
\scalebox{0.75}{
\resizebox{\textwidth}{!}{%
\begin{tabular}{llrrrrrr}
\toprule
 &  & \multicolumn{3}{c}{CREAK} & \multicolumn{3}{c}{e-SNLI} \\ \cline{3-8} 
 &  & \multicolumn{1}{l}{Standard} & \multicolumn{1}{l}{Explain} & \multicolumn{1}{l}{Permute} & \multicolumn{1}{l}{Standard} & \multicolumn{1}{l}{Explain} & \multicolumn{1}{l}{Permute} \\ \hline
\multirow{5}{*}{
\vspace{-2cm}
\begin{tabular}[c]{@{}l@{}}Accuracy\\ ($\delta_{acc}$)\end{tabular}} & No Cue & 84.2 & 85.0 & \textbf{86.2} & \textbf{91.6} & 89.2 & 90.0 \\ \cline{2-8} 
& Sentence Length & \begin{tabular}[c]{@{}r@{}}60.4\\ \small (-23.8)\end{tabular} & \textbf{\begin{tabular}[c]{@{}r@{}}80.2\\ \small (-4.8)\end{tabular}} & \begin{tabular}[c]{@{}r@{}}67.6\\ \small (-18.6)\end{tabular} & \begin{tabular}[c]{@{}r@{}}69.8\\ \small (-21.8)\end{tabular} & \textbf{\begin{tabular}[c]{@{}r@{}}76.2\\ \small (-13.0)\end{tabular}} & \begin{tabular}[c]{@{}r@{}}72.2\\ \small (-17.8)\end{tabular} \\ \cline{2-8} 
& Present Tense & \begin{tabular}[c]{@{}r@{}}74.6\\ \small (-9.6)\end{tabular} & \textbf{\begin{tabular}[c]{@{}r@{}}80.2\\ \small (-4.8)\end{tabular}} & \begin{tabular}[c]{@{}r@{}}75.4\\ \small (-10.8)\end{tabular} & \begin{tabular}[c]{@{}r@{}}85.8\\ \small (-5.8)\end{tabular} & \textbf{\begin{tabular}[c]{@{}r@{}}88.0\\ \small (-1.2)\end{tabular}} & \begin{tabular}[c]{@{}r@{}}80.2\\ \small (-9.8)\end{tabular} \\ \cline{2-8} 
& Embedding Cluster & \begin{tabular}[c]{@{}r@{}}69.2\\ \small (-15.0)\end{tabular} & \textbf{\begin{tabular}[c]{@{}r@{}}78.6\\ \small (-6.4)\end{tabular}} & \begin{tabular}[c]{@{}r@{}}74.8\\ \small (-11.4)\end{tabular} & \begin{tabular}[c]{@{}r@{}}70.6\\ \small (-21.0)\end{tabular} & \textbf{\begin{tabular}[c]{@{}r@{}}88.6\\ \small (-0.6)\end{tabular}} & \begin{tabular}[c]{@{}r@{}}77.4\\ \small (-12.6)\end{tabular} \\ \cline{2-8} 
& Average & \begin{tabular}[c]{@{}r@{}}68.1\\ \small (-16.1)\end{tabular} & \textbf{\begin{tabular}[c]{@{}r@{}}79.7\\ \small (-5.3)\end{tabular}} & \begin{tabular}[c]{@{}r@{}}72.6\\ \small (-13.6)\end{tabular} & \begin{tabular}[c]{@{}r@{}}75.4\\ \small (-16.2)\end{tabular} & \textbf{\begin{tabular}[c]{@{}r@{}}84.3\\ \small (-4.9)\end{tabular}} & \begin{tabular}[c]{@{}r@{}}76.6\\ \small (-13.4)\end{tabular} \\
\hline
\multirow{4}{*}{\begin{tabular}[c]{@{}l@{}}Prediction-\\ Feature \\ Correlation\end{tabular}} & Sentence Length & 0.847 & \textbf{0.325} & 0.457 & 0.467 & \textbf{0.291} & 0.382 \\ 
& Present Tense & 0.305 & \textbf{0.146} & 0.319 & 0.217 & \textbf{0.143} & 0.322 \\
& Embedding Cluster & 0.563 & \textbf{0.288} & -0.427 & 0.595 & \textbf{0.141} & -0.303 \\
& Average & 0.572 & 0.253 & \textbf{0.116} & 0.426 & 0.192 & \textbf{0.134} \\ \bottomrule
\end{tabular}
}}
\caption{Results on CREAK and e-SNLI when explanations are permuted to be completely irrelevant to the input, in comparison with standard finetuning and explanation-based finetuning (with valid explanations).}
\vspace{-0.1in}
\label{tab: explanation-quality}
\end{table*}

%% file: tables/bootstrapping_exps.tex
\begin{figure*}
  \centering
  
  \begin{subfigure}[b]{0.48\textwidth}
  \pgfplotsset{width=0.95\columnwidth, height=3.5cm}
    \centering
    \begin{tikzpicture}  
        \begin{axis}
        [  
            ybar,
            ymin=50, ymax=100,
            ytick={50,60,70,80,90,100},
            major x tick style = transparent,
            bar width=4pt,
            enlarge x limits=0.25,
            ylabel={Accuracy},
            ylabel style={font=\scriptsize},
            symbolic x coords={Sentence Length, Present Tense, Embedding Cluster, Plural Noun, Average},  
            xtick=data,
            xticklabel style={font=\scriptsize, rotate=30, anchor=east}, 
            yticklabel style={font=\scriptsize},
            axis x line*=bottom,
            axis y line*=left,
            legend style={draw=none},
            legend cell align=left,
            legend style={
                    at={(1.0,1.05)},
                    anchor=south east,
                    column sep=1ex,
                    font=\tiny,
                    legend columns=3
            },
        ]
        \addplot[ybar,  fill=Color1, postaction={}] coordinates {
            (Sentence Length, 60.4) (Present Tense, 74.6) (Embedding Cluster, 69.2) (Plural Noun, 72.2) (Average, 69.1)
        };
        \addplot[ybar,  fill=Color2, postaction={pattern=north west lines}] coordinates {
            (Sentence Length, 80.2) (Present Tense, 80.2) (Embedding Cluster, 78.6) (Plural Noun, 77.2) (Average, 79.1)
        };  
        \addplot[ybar, fill=Color3, postaction={pattern=north east lines}] coordinates {
            (Sentence Length, 78.6) (Present Tense, 77.2) (Embedding Cluster, 73.0) (Plural Noun, 80.8) (Average, 77.4)
        };  
        \legend{Standard, Explain (Human), Explain (Bootstrap)}  
        \end{axis}  
    \end{tikzpicture}
    \vspace{-0.1in}
    \caption{Accuracy($\uparrow$) on CREAK}
        \vspace{-0.1in}

    \label{fig:creak_accuracy_plot}
  \end{subfigure}
  \hfill 
  \begin{subfigure}[b]{0.48\textwidth}
\pgfplotsset{width=0.95\columnwidth, height=3.5cm}
    \centering
    \begin{tikzpicture}  
        \begin{axis}
        [  
            ybar,
            ymin=60, ymax=100,
            ytick={60,70,80,90,100},
            major x tick style = transparent,
            bar width=4pt,
            enlarge x limits=0.25,
            symbolic x coords={Sentence Length, Present Tense, Embedding Cluster, Plural Noun},  
            xtick=data,
            xticklabel style={font=\scriptsize, rotate=30, anchor=east},  
            yticklabel style={font=\scriptsize},
            axis x line*=bottom,
            axis y line*=left
        ]
        \addplot[ybar,  fill=Color1, postaction={}] coordinates {
            (Sentence Length, 69.8) (Present Tense, 76.0) (Embedding Cluster, 70.6) (Plural Noun, 69.0)
        };
        \addplot[ybar,  fill=Color2, postaction={pattern=north west lines}] coordinates {
            (Sentence Length, 76.2) (Present Tense, 86.6) (Embedding Cluster, 89.2) (Plural Noun, 85.4)
        };  
        \addplot[ybar, fill=Color3, postaction={pattern=north east lines}] coordinates {
            (Sentence Length, 76.0) (Present Tense, 87.0) (Embedding Cluster, 79.2) (Plural Noun, 83.4)
        };  
        \end{axis}  
    \end{tikzpicture}
    \vspace{-0.1in}
    \caption{
    Accuracy($\uparrow$) on e-SNLI
    }
        \vspace{-0.1in}
    \label{fig:esnli_accuracy_plot}
  \end{subfigure}

  \begin{subfigure}[b]{0.48\textwidth}
\pgfplotsset{width=0.95\columnwidth, height=3.5cm}
    \centering
    \begin{tikzpicture}  
        \begin{axis}
        [  
            ybar,
            ymin=0, ymax=1,
            ytick={0, 0.2, 0.4, 0.6, 0.8, 1},
            major x tick style = transparent,
            bar width=4pt,
            enlarge x limits=0.25,
            ylabel={Prediction-Feature Correlation},
            ylabel style={font=\scriptsize},
            symbolic x coords={Sentence Length, Present Tense, Embedding Cluster, Plural Noun, Average},  
            xtick=data,
            xticklabel style={font=\scriptsize, rotate=30, anchor=east},  
            yticklabel style={font=\scriptsize},
            axis x line*=bottom,
            axis y line*=left,
            legend style={draw=none},
            legend cell align=left,
        ]
        \addplot[ybar,  fill=Color1, postaction={}] coordinates {
            (Sentence Length, 0.847) (Present Tense, 0.305) (Embedding Cluster, 0.563) (Plural Noun, 0.445) (Average, 0.540)
        };
        \addplot[ybar,  fill=Color2, postaction={pattern=north west lines}] coordinates {
            (Sentence Length, 0.325) (Present Tense, 0.146) (Embedding Cluster, 0.288) (Plural Noun, 0.170) (Average, 0.232)
        };  
        \addplot[ybar, fill=Color3, postaction={pattern=north east lines}] coordinates {
            (Sentence Length, 0.045) (Present Tense, 0.167) (Embedding Cluster, 0.429) (Plural Noun, 0.129) (Average, 0.193)
        };  
        \end{axis}  
    \end{tikzpicture}
    \vspace{-0.1in}
    \caption{
    Prediction-Feature Correlation($\downarrow$) on CREAK
    }
    \vspace{-0.05in}
    \label{fig:creak_correlation_plot}
  \end{subfigure}
  \hfill 
  \begin{subfigure}[b]{0.48\textwidth}
\pgfplotsset{width=0.95\columnwidth, height=3.5cm}
    \centering
    \begin{tikzpicture}  
        \begin{axis}
        [  
            ybar,
            ymin=0, ymax=1.0,
            ytick={0, 0.2, 0.4, 0.6,0.8,1.0},
            major x tick style = transparent,
            bar width=4pt,
            enlarge x limits=0.25,
            symbolic x coords={Sentence Length, Present Tense, Embedding Cluster, Plural Noun},  
            xtick=data,
            xticklabel style={font=\scriptsize, rotate=30, anchor=east},  
            yticklabel style={font=\scriptsize},
            axis x line*=bottom,
            axis y line*=left,
            legend style={draw=none},
            legend cell align=left,
            legend style={
                    at={(1.0,1.05)},
                    anchor=south east,
                    column sep=1ex,
                    font=\tiny,
                    legend columns=3
            },
        ]
        \addplot[ybar,  fill=Color1, postaction={}] coordinates {
            (Sentence Length, 0.467) (Present Tense, 0.336) (Embedding Cluster, 0.595) (Plural Noun, 0.578)
        };
        \addplot[ybar,  fill=Color2, postaction={pattern=north west lines}] coordinates {
            (Sentence Length, 0.291) (Present Tense, 0.055) (Embedding Cluster, 0.147) (Plural Noun, 0.221)
        };  
        \addplot[ybar, fill=Color3, postaction={pattern=north east lines}] coordinates {
            (Sentence Length, 0.343) (Present Tense, 0.133) (Embedding Cluster, 0.350) (Plural Noun, 0.257)
        };  
        \end{axis}  
    \end{tikzpicture}
    \vspace{-0.1in}
    \caption{
    Prediction-Feature Correlation($\downarrow$) on e-SNLI
    }
    \vspace{-0.05in}
    \label{fig:esnli_correlation_plot}
  \end{subfigure}
  \hfill
  \caption{Results for finetuning with bootstrapped explanations (\textbf{Explain (Bootstrap)}), in comparison to finetuning without explanations (\textbf{Standard}) and finetuning with human-written explanations (\textbf{Explain (Human)}).}
  \label{fig:bootstrap}
\end{figure*}
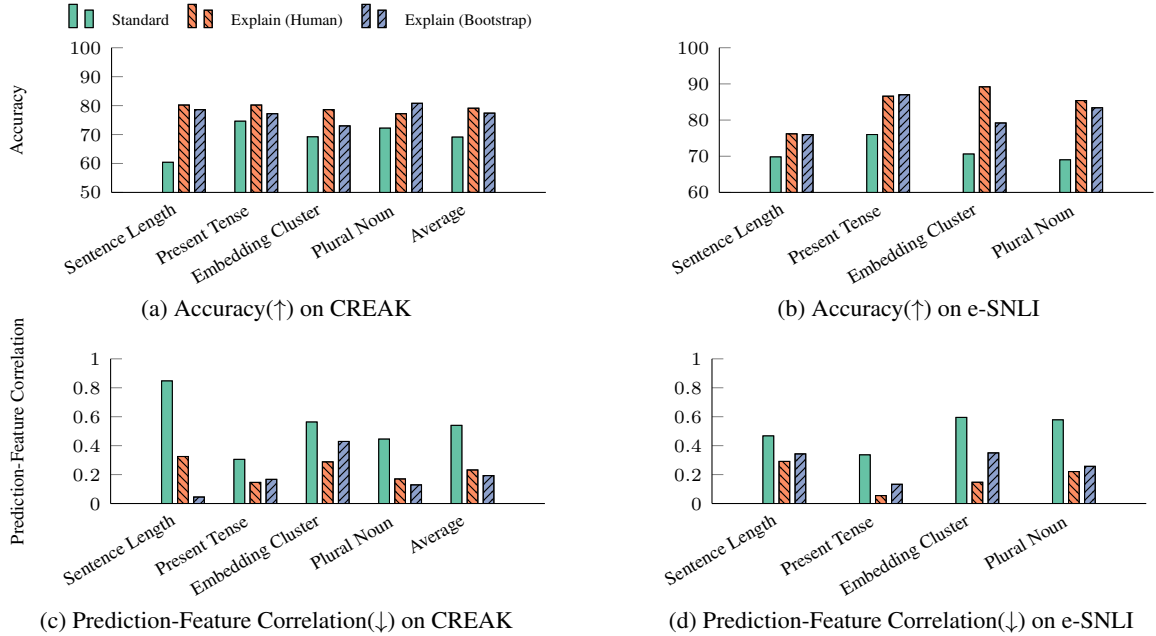

%% file: templates/9-Conclusion.tex
\vspace{-0.1in}
\section{Conclusion}

We propose explanation-based finetuning, a general method for reducing model reliance on spurious cues present in the training data. Specifically, in addition to predicting a label, models are finetuned to also generate a free-text explanation in support of its prediction. We perform experiments  on a diverse set of classification tasks involving different types of spurious features.
Results show that our method makes the models substantially more robust towards spurious features, as measured by both accuracy and correlation-based metrics. The efficacy of our method generalizes to different model sizes and families, though larger models tend to benefit more. Moreover, we observe that the stronger the spurious correlation in the data, the more helpful our method is. Interestingly, we show that highly relevant explanations are not absolutely necessary, since permuted explanations still provide around 25\% of the accuracy benefits observed with non-permuted explanations. What is most notable is that even with model-generated explanations, our method works almost as well as with human-written ones, implying its potential applicability to the vast majority of datasets for which human-written explanations are not available.

%% file: templates/10-Limitations.tex
\section*{Limitations}

We notice a few key limitations of our approach. Similar to what was shown by previous interpretability studies \cite[][i.a.]{camburu2018snli}, incorporating explanations comes with some penalty on in-distribution accuracy when there is no spurious cue. This penalty decreases as model size increases, potentially because it is less challenging for larger models to generate good explanations. The second limitation is that our artificially constructed training set
may not reflect the strength of the studied spurious cues in the real world. In our main experiments, we focus on the case where one spurious cue is perfectly correlated with the target label. For further exploration, we can study the alternative setting where there are multiple weak spurious cues instead of a single strong one. Finally, our work here is limited  by the scope of the experiments. We only experiment with generative LMs and binary classification tasks. Also, because of resource constraints, we only consider four datasets and eight types of spurious cues (including dataset-independent and dataset-specific ones). Additional experiments using a wider variety of spurious cues and datasets would help to shed light on how our method generalizes to other scenarios.

%% file: templates/11-Ethical_Considerations.tex
\section*{Ethics Statement}

\minisection{Potential risks} 
While our work on overcoming spurious cues is related to the idea of debiasing models, it is important to note that our results do not indicate that our method is the best to tackle socially harmful biases against marginalized groups, like gender or racial biases. We have not run any experiments following this direction, and it is important to make this distinction so that the reader does not misunderstand the goal of this paper.

\minisection{Intended Use} 
Our models and methods shown here are for research purposes only. They should not be deployed in the real world as solutions without further evaluation.

%% file: templates/Appendix.tex
\newpage

\appendix

\section{Extended Results}

\subsection{Results Under ``No Cue'' Condition}

\input{tables/baseline_table.tex}

Under the ``no cue'' condition (i.e., when the training set is unskewed), we report the test accuracy of GPT-3 (Davinci) under finetuning (n=1,000), few-shot (n=10), and zero-shot settings. Results are shown in Table~\ref{tab: baselines}. 
Across the four different datasets, the model finetuned on 1,000 examples achieves much higher accuracies compared to 10-shot or zero-shot prompting.

Comparing standard finetuning and explanation-based finetuning, across all these experiments, we only find an obvious increase (+6.7) on CREAK under the few-shot setting and a slight increase (+0.4) on ComVE under the zero-shot setting. In all other cases, the accuracy either drops or stays the same.

\subsection{Results for Other Models}
\label{appendix: other models}
\input{tables/ada_table}
\input{tables/babbage_table}
\input{tables/curie_table}
\input{tables/t5_table.tex}
\input{tables/bart_table.tex}
\input{tables/opt_table.tex}

In our main experiments in Section~\ref{sec: results} and Section~\ref{sec: analysis}, we use OpenAI GPT-3 (Davinci(175B), Curie(13B), Babbage(6.7B), and Ada (2.7B), since their relatively large size may allow for the generation of higher-quality experiments, as suggested by \cite{wei2022chain}. 

We also generalize this approach to other model families including T5-base (220M), BART-base (110M), and OPT (1.3B).
Table~\ref{tab: t5-base} and Table~\ref{tab: bart-base} show the results for these T5 and BART models respectively. 
Under the ``no cue'' condition, their performance is generally much worse than GPT-3 models. The penalty of introducing explanations in finetuning is also more striking, oftentimes resulting in an accuracy around or lower than chance (50.0).
When the training set contains spurious cues, our method still generally works for both T5 and BART on three of the four datasets, as measured by $\delta_{acc}^f(M, FT)$ and $corr_f(M_f^{FT})$. However, the absolute accuracy is almost consistently lower for explanation-based finetuning than for standard finetuning, most likely due to the huge penalty under the ``no cue'' condition in the first place.

As an exception, on the SBIC dataset, our method does not always work well. For the T5 model, across all spurious features, explanation-based finetuning results in a similar or worse $\delta_{acc}$ (the difference is always less than 2.0 percent). It also fails to reduce the prediction-feature correlation for any spurious feature except the ``embedding cluster'' one, where the correlation only decreases by 0.03.
For the BART model, our method does make it more robust to the ``embedding cluster'' and the ``plural noun'' cues but no other cues, as reflected by both the accuracy drop and the prediction-feature correlation. We hypothesize that this is because of the model does not rely heavily on the cues in the first place, as shown by the lower prediction-feature correlations in the case of standard finetuning. This reconfirms our observation from Section~\ref{sec:discussion}.

We further generalize our method to OPT (1.3b) with results shown in Table~\ref{tab: opt results}. Its performance under the ``no cue'' condition is comparable with the performance of Ada (Table~\ref{tab: ada results}).
Compared to standard finetuning, our method effectively mitigates the accuracy drop ($\delta_{acc}^f(M, FT)$) and the correlation between the prediction and the cue ($corr_f(M_f^{FT})$) averaged across all datasets. These results are mixed across cues however: the absolute accuracies of the with-explanation models for most tasks are lower when the ``present tense'' cue is introduced but are improved for all tasks in case of the ``embedding cluster'' cue.

Generally, compared to GPT-3, our method still works on most of the datasets for T5 and BART, but with smaller benefits. This is most likely because explanation generation is in itself a challenging task for smaller models, thus resulting in a larger penalty on accuracy in the ``no cue'' condition. 
The results of the larger OPT model lend greater credence to the validity of the assumption.

\section{Extended Analysis}

\subsection{Does knowledge of the cue improve model robustness via few-shot prompting?}
\input{tables/fewshot_table.tex}

In our main experiments, we only consider datasets that come with human-annotated explanations for all training instances. However, this is untrue for the vast majority of datasets in the real world. Here, we want to explore if it is possible to overcome the cue \textit{without} large-scale human-written explanations available. Specifically, given only a few examples of human-written explanations, can we still make the model more robust, if we have knowledge about what the spurious feature is?

Specifically, we take standard-finetuned models trained on the skewed training sets. Then, we use 10 training examples to construct the few-shot prompt. In the standard prompting setting, we only include the input and the label; for explanation-based prompting, we additionally include a free-text explanation before the label. For both settings, the set of few-shot examples is randomly shuffled and unskewed (i.e., they do not exhibit the spurious correlation).

We experiment with e-SNLI in this analysis.
The results, as shown in Table~\ref{tab: few-shot-over-finetuned experiments}, indicate that for syntactic spurious cues, standard prompting significantly helps the standard model become more robust to them. 
The correlations between the model predictions and the spurious cue drop by 0.297 to 0.359 for the three syntactic spurious cues.
However, there is no evidence that few-shot prompting benefits when the ``embedding cluster'' cue is introduced. 
Although adding explanations is shown to be effective in finetuning, it does not help as much in few-shot prompting, in terms of either accuracy or prediction-feature correlation.

\subsection{Increasing the number of finetuning examples from 1k to 4k}

In this analysis, we examine the effect of increasing the number of training examples for finetuning from 1k to 4k. This is to investigate the hypothesis that increasing the number of training examples will make it easier for models to learn, and subsequently overfit on the spurious cue.

\minisection{e-SNLI Experiments.} We repeat the experiments used to create Figure~\ref{figure: finetune-n figure} with the modification that instead of being trained on 1k examples, models are trained on 4k examples. These results are shown in Table~\ref{tab: n-finetune-bias-strength}. In the table, we find that the accuracies of both the standard and finetuned models improve when we increase the number of training examples. The average standard finetuning model increases by 2.3 while for the explanation-based finetuned models this increase is 5.2. Correspondingly, the average accuracy gap also increases between the standard and explanation-based models from 4.52 in the n=1k to 6.70 (+2.18).

Looking at the prediction-feature correlation, we note that the average correlation does not change substantially for both the standard finetuning and explanation-based finetuning after increasing the number of training examples to 4k.

Overall, these results provide evidence that having an increased number of examples tends to benefit both standard and explanation based finetunes with explanation-based finetunes being able to benefit more. However, in the case that the training set correlation between the target label and the spurious cue is 1.0, we note that the performance for the standard finetuning drops substantially.

\input{tables/esnli_cue_strength_table.tex}

\minisection{ComVE and SBIC Experiments.} Furthering the results from the previous analysis, we investigate the effect of increasing the number of finetuning examples in the cases where we found the effect of explanation-based finetuning to be the weakest in Table~\ref{tab: davinci results}. Specifically, we investigate SBIC and ComVE under the present tense and sentence length spurious cues by rerunning the experiments under this setting with the modification of increasing the training set size from 1k to 4k. These results are shown in Table~\ref{tab: n-increase experiments}.

\input{tables/sbic_comve_finetune_n.tex}

These results provide strong confirmation that increasing the number of examples when the spurious cue is perfectly correlated with the label substantially degrades model performance. Under the setting where we only have 1k training examples, the average accuracy difference between standard and explanation-based finetuning across both cues and datasets is 1.0 in favor of standard finetuning. This difference when we have 4k training examples is 8.4 in favor of explanation-based finetuning. It is worth noting that in three out of the four settings in this experiment (everything except length bias for SBIC), in the n=1k setting, standard finetuning does not provide a benefit. However, if we increase n to 4k, that increases the model's susceptibility to the cue enough that explanation-based finetuning outperforms standard finetuning, a reversal of the original results.

\subsection{Dataset-Specific Spurious Cues}
\label{appendix: sec: domain cues}

\input{tables/domain_sepcific_davinci_table.tex}

In addition to the four common spurious cues in the main text, we also construct dataset-specific spurious correlations to simulate realistic cues that can naturally appear in each dataset:

\minisection{Higher Perplexity (CREAK).}
    Using GPT-2 to measure perplexity, we filter the data into a set with above-median perplexity and a set with below-median perplexity. This feature is considered to be present if the perplexity of the sentence is higher than the median perplexity and is positively labeled.

\minisection{Gender Female (e-SNLI).}
    If the premise contains female-related pronouns (woman, women, girl, lady, etc.), we consider the ``gender female'' spurious cue to be present. The aforementioned words frequently appear in the e-SNLI dataset when the sentence is relevant to females.
    
\minisection{Username Mentions (SBIC).}
    If the social media post contains an ``@'' sign, meaning the author might be tagging or directly replying to other users on social media, we consider the spurious cue to be present. 
    This feature is supposed to have no causal relationship with whether a post is offensive.

\minisection{POS-tag of Swapped Word (ComVE).} 
    The ComVE dataset requires us to compare two sentences and output which sentence makes more sense, the two sentences have high lexical overlaps. We consider the part of speech (POS) of the first word which is different between the two sentences and say that the POS tag of swapped word spurious cue is present if this word is a noun.

Table~\ref{tab: domain} shows the performance of GPT-3 (Davinci). When adding ``gender female'' spurious cues to the e-SNLI dataset, we find strong evidence that explanations make the model less susceptible to the spurious cue. In standard finetuning, the prediction-feature correlation is 0.684 and the accuracy is 55.8, suggesting the model relies heavily on the spurious pattern. 
Meanwhile, for the model finetuned with explanations, this correlation drops to 0.080, and the accuracy increases to 86.6. The results for dataset-specific cues of the ComVE and CREAK datasets are consistent with our finding that our approach is most effective when the spurious cues highly impact the model performance. On the SBIC dataset, explanation-based finetuning only decreases the prediction-feature correlation by $0.076$. This could be due to the fact that the ``username mention'' cue is the most shallow one among all domain-specific cues, since the model only needs to detect one token (``@''), which makes it surprisingly easy for it to pick up the cue.

\section{Implementation Details}
\label{appendix:sec:implemetation_details}
\subsection{Spurious Cue Implementation}
The implementation of the ``present tense'' and ``plural noun'' spurious cues described in Section~\ref{sec: induced bias} and the ``POS-tag of swapped word'' cue in the Section~\ref{appendix: sec: domain cues} involve tokenizing and performing POS tagging on the inputs. The tokenizer and POS-tagger we use are implemented by \cite{bird2009natural} in the NLTK toolkit \footnote{\url{https://www.nltk.org/}}.

For the ``higher perplexity'' spurious cue for the CREAK dataset, we compute the GPT-2 perplexity of the input text using the metric module implemented in the Huggingface Evaluate package
\footnote{\url{https://github.com/huggingface/evaluate}}. Its license is Apache License 2.0.

\subsection{Models and Hyperparameters}

All our code are attached as the supplemental materials.

\minisection{OpenAI Models} 
We finetuned GPT-3 \cite{NEURIPS2020_1457c0d6} from OpenAI's standard API\footnote{\url{https://beta.openai.com/docs/api-reference}} in different sizes (\texttt{Davinci} and \texttt{Ada}). Its license is MIT license.
The GPT-3 models are finetuned for four epochs (default setting on the OpenAI API), and the other hyperparameters (e.g. learning rates) are the default values. with the exception of the models trained with 4k examples which were only trained for one epoch with an increased learning rate (0.2) to reduce costs.

\minisection{Huggingface Models}
T5 \cite{raffel2020exploring}, BART \cite{lewis2019bart}, and OPT \cite{zhang2022opt} are implemented with HuggingFace Transformers\footnote{\url{https://github.com/huggingface/transformers}}. 
The pretrained model checkpoints we use are the \texttt{t5-base} (220M parameters), \texttt{facebook/bart-base} (110M parameters) and \texttt{facebook/opt-1.3b} (1.3B parameters). Their licenses are Apache License 2.0 (T5 and BART) or other\footnote{\url{https://huggingface.co/facebook/opt-1.3b/blame/aa6ac1e23bb9a499be2b7400079cd2a7b8a1309a/LICENSE.md}} (OPT).
We use the \texttt{conditional generation} classes for T5 \footnote{\url{https://huggingface.co/docs/transformers/model_doc/t5\#transformers.T5ForConditionalGeneration}} and BART \footnote{\url{https://huggingface.co/docs/transformers/model_doc/bart\#transformers.BartForConditionalGeneration}}, and use the \texttt{auto model for causalLM} class for OPT \footnote{\url{https://huggingface.co/docs/transformers/model_doc/auto\#transformers.AutoModelForCausalLM}}
from Huggingface to finetune the pretrained models.
To remain consistent with the finetuning of OpenAI models, the T5 and BART models are finetuned with 1,000 training examples and run for 4 training epochs. 
The batch size is set to 8 and the learning rate is set to 2e-5 with the max sequence length being 128.
The OPT model may take longer to converge, we consistently use 1,000 training examples and set batch size to 8, but the standard finetuning on CREAK, and the with-explanation finetuning on e-SNLI and ComVE run for six epochs, the learning rate of the standard finetuning on CREAK and SBIC, and the with-explanation finetuning on ComVE is set to 1e-5, the learning rate of the with-explanation finetuning on  SBIC is set to 6e-5. For other settings, the number of training epochs is set to 4 and the learning rate is set to 2e-5.
Our finetuning experiments of T5 and BART are run on a Kepler K80 GPU. 
The finetuning of the OPT models is run on an RTX A6000.
Each finetuning takes 5 to 10 minutes depending on the task.

\subsection{Computational Resources}

All experiments performed using GPT-3 including all finetuning were performed using the OpenAI public API. We note that every finetuning experiment on each cue and dataset in this paper costs around $\$10$ to perform. Across all our datasets, creating a finetuned model involving 1k samples cost around $\$5$ when tuned without explanations and $\$7$ with explanations. Performing evaluation with these finetuned models then cost around a dollar when evaluating on 500 samples.

All other experiments involving heavy computational resources such as finetuning T5 and BART were performed on Google Colaboratory with GPU-accelerated notebooks available on the pro subscription.

\section{Datasets Details}

\subsection{Dataset URLs and Licenses}

Listed below are all the details and licenses (where available) for the datasets used in this paper. All datasets used were research datasets and used for their intended purposes. None of the data used in this paper contains any sensitive information. A disclaimer has been added at the start of this paper for offensive content given that the SBIC dataset contains examples of hate speech.

\minisection{CREAK \cite{onoe2021creak}}: \url{https://github.com/yasumasaonoe/creak}

\minisection{e-SNLI \cite{camburu2018snli}}: \url{https://github.com/OanaMariaCamburu/e-SNLI}, license: MIT License, \url{https://github.com/OanaMariaCamburu/e-SNLI/blob/master/LICENSE}

\minisection{ComVE \cite{wang2019does}}:\\ \url{https://github.com/wangcunxiang/SemEval2020-Task4-Commonsense-Validation\\-and-Explanation}, license: CC BY 4.0 

\minisection{SBIC \cite{sap2019social}}: \url{https://maartensap.com/social-bias-frames/SBIC.v2.tgz}, license: CC BY 4.0

\subsection{Label-Feature Correlation in Unskewed Training Sets}
\label{appendix:base_correlations}

\input{tables/corr_base_table.tex}

The correlation between the ground-truth label and the spurious cues on the randomly selected 1,000 training sets is shown in Table~\ref{tab: corr base}.
There are no artificially introduced spurious correlations in this training set. According to the correlations in the table, we claim that the ``no cue'' training set is unskewed, except for the ``embedding cluster'' on the SBIC dataset where this correlation is $0.378$, implying that the embedding vectors for the offensive social media posts are clustered together.

\section{Sample Outputs}
Here are some randomly selected sample outputs of the Davinci model for the CREAK dataset, with standard-finetuning and explanation-based finetuning.

\input{tables/standard_sample_outputs}
\input{tables/explain_sample_outputs}

%% file: tables/baseline_table.tex
\begin{table*}[t]
\centering
\scalebox{0.8}{
\resizebox{\textwidth}{!}{%
\begin{tabular}{@{}lllllllll@{}}
\toprule
\multirow{2}{*}{} & \multicolumn{2}{c}{ComVE} & \multicolumn{2}{c}{CREAK} & \multicolumn{2}{c}{e-SNLI} & \multicolumn{2}{c}{SBIC} \\ \cmidrule(l){2-9} 
 & Standard & Explain & Standard & Explain & Standard & Explain & Standard & Explain \\ \midrule
\begin{tabular}[c]{@{}l@{}}Finetuned\\ (n=1k)\end{tabular} & \textbf{97.0} & 95.5 & 84.2 & \textbf{85.0} & \textbf{91.6} & 89.2 & \textbf{79.0} & 75.0 \\ \midrule
\begin{tabular}[c]{@{}l@{}}Fewshot\\ (n=10)\end{tabular} & \textbf{54.0} & \textbf{54.0} & 67.5 & \textbf{74.0} & \textbf{59.0} & 55.5 & \textbf{72.0} & 66.0 \\ \midrule
Zero-Shot & 47.6 & \textbf{48.0} & \textbf{57.0} & 55.5 & \textbf{51.4} & 50.6 & 56.6 & \textbf{62.8} \\ \bottomrule
\end{tabular}%
}}
\caption{Accuracies under the ``no cue'' condition for all datasets across different finetuning and prompting strategies.}
\label{tab: baselines}
\end{table*}

%% file: tables/ada_table.tex
\begin{table*}[t]
\centering
\resizebox{\textwidth}{!}{%
\begin{tabular}{llrrrrrrrr}
    \toprule
     &  & \multicolumn{2}{c}{ComVE} & \multicolumn{2}{c}{CREAK} & \multicolumn{2}{c}{e-SNLI} & \multicolumn{2}{c}{SBIC} \\ \cline{3-10} 
     &  & Standard & Explain & Standard & Explain & Standard & Explain & Standard & Explain \\ \hline
    \multirow{6}{*}{
    \vspace{-2cm} 
    \begin{tabular}[c]{@{}l@{}}Accuracy\\ ($\delta_{acc}$)\end{tabular}
    } & No Cue & \textbf{79.2} & 52.4 & \textbf{71.6} & 62.6 & \textbf{88.0} & 76.4 & \textbf{80.0} & 74.6 \\ \cline{2-10} 
     & Sentence Length & \begin{tabular}[c]{@{}r@{}}44.8\\ \small (-34.4)\end{tabular} & \textbf{\begin{tabular}[c]{@{}r@{}}48.4\\ \small (-4.0)\end{tabular}} & \begin{tabular}[c]{@{}r@{}}53.0\\ \small (-18.6)\end{tabular} & \textbf{\begin{tabular}[c]{@{}r@{}}56.6\\ \small (-6.0)\end{tabular}} & \begin{tabular}[c]{@{}r@{}}60.4\\ \small (-27.6)\end{tabular} & \textbf{\begin{tabular}[c]{@{}r@{}}64.6\\ \small (-11.8)\end{tabular}} & \begin{tabular}[c]{@{}r@{}}\textbf{53.6}\\ \small (-26.4)\end{tabular} & \begin{tabular}[c]{@{}r@{}}49.6\\ \small \textbf{(-25.0)}\end{tabular} \\ \cline{2-10} 
     & Present Tense & \begin{tabular}[c]{@{}r@{}}53.2\\ \small (-26.0)\end{tabular} & \textbf{\begin{tabular}[c]{@{}r@{}}54.0\\ \small (1.6)\end{tabular}} & \begin{tabular}[c]{@{}r@{}}55.2\\ \small (-16.4)\end{tabular} & \textbf{\begin{tabular}[c]{@{}r@{}}55.8\\ \small (-6.8)\end{tabular}} & \begin{tabular}[c]{@{}r@{}}67.4\\ \small (-20.6)\end{tabular} & \textbf{\begin{tabular}[c]{@{}r@{}}69.6\\ \small (-6.8)\end{tabular}} & \begin{tabular}[c]{@{}r@{}}70.6\\ \small (-9.4)\end{tabular} & \textbf{\begin{tabular}[c]{@{}r@{}}75.2\\ \small (0.6)\end{tabular}} \\ \cline{2-10} 
     & Embedding Cluster & \begin{tabular}[c]{@{}r@{}}47.6\\ \small (-31.6)\end{tabular} & \textbf{\begin{tabular}[c]{@{}r@{}}48.4\\ \small (-4.0)\end{tabular}} & \begin{tabular}[c]{@{}r@{}}50.2\\ \small (-21.4)\end{tabular} & \textbf{\begin{tabular}[c]{@{}r@{}}51.8\\ \small (-10.8)\end{tabular}} & \begin{tabular}[c]{@{}r@{}}55.8\\ \small (-32.2)\end{tabular} & \textbf{\begin{tabular}[c]{@{}r@{}}58.0\\ \small (-18.4)\end{tabular}} & \begin{tabular}[c]{@{}r@{}}\textbf{56.0}\\ \small (-24.0)\end{tabular} & \begin{tabular}[c]{@{}r@{}}55.0\\ \small \textbf{(-19.6)}\end{tabular} \\ \cline{2-10} 
     & Plural Noun & \begin{tabular}[c]{@{}r@{}}51.8\\ \small (-27.4)\end{tabular} & \textbf{\begin{tabular}[c]{@{}r@{}}53.8\\ \small (1.4)\end{tabular}} & \begin{tabular}[c]{@{}r@{}}53.0\\ \small (-18.6)\end{tabular} & \textbf{\begin{tabular}[c]{@{}r@{}}53.8\\ \small (-8.8)\end{tabular}} & \begin{tabular}[c]{@{}r@{}}52.6\\ \small (-35.4)\end{tabular} & \textbf{\begin{tabular}[c]{@{}r@{}}58.4\\ \small (-18.0)\end{tabular}} & \begin{tabular}[c]{@{}r@{}}70.8\\ \small (-9.2)\end{tabular} & \textbf{\begin{tabular}[c]{@{}r@{}}71.8\\ \small (-2.8)\end{tabular}} \\ \cline{2-10} 
     & Average & \begin{tabular}[c]{@{}r@{}}49.4\\ \small (-29.9)\end{tabular} & \begin{tabular}[c]{@{}r@{}}\textbf{51.2}\\ \small \textbf{(-1.3)}\end{tabular} & \begin{tabular}[c]{@{}r@{}}52.9\\ \small (-18.8)\end{tabular} & \begin{tabular}[c]{@{}r@{}}\textbf{54.5}\\ \small \textbf{(-8.1)}\end{tabular} & \begin{tabular}[c]{@{}r@{}}59.1\\ \small (-29.0)\end{tabular} & \textbf{\begin{tabular}[c]{@{}r@{}}62.7\\ \small (-13.8)\end{tabular}} & \begin{tabular}[c]{@{}r@{}}62.8\\ \small (-17.3)\end{tabular} & \begin{tabular}[c]{@{}r@{}}\textbf{62.9}\\ \small \textbf{(-11.7)}\end{tabular} \\ \hline
    \multirow{5}{*}{\begin{tabular}[c]{@{}l@{}}Correlation between \\ Model's Prediction\\ and Spurious Feature\end{tabular}} & Sentence Length & 0.870 & \textbf{0.778} & 0.847 & \textbf{0.590} & 0.644 & \textbf{0.531} & \textbf{0.676} & 0.712 \\
     & Present Tense & 0.956 & \textbf{0.948} & 0.738 & \textbf{0.573} & 0.586 & \textbf{0.408} & 0.461 & \textbf{0.258} \\ 
     & Embedding Cluster & 0.858 & \textbf{0.807} & 0.751 & \textbf{0.705} & 0.876 & \textbf{0.753} & 0.447 & \textbf{0.428} \\ 
     & Plural Noun & 0.853 & \textbf{0.774} & 0.775 & \textbf{0.484} & 0.911 & \textbf{0.702} & 0.393 & \textbf{0.234} \\ 
     & Average & 0.884 & \textbf{0.827} & 0.778 & \textbf{0.588} & 0.754 & \textbf{0.599} & 0.494 & \textbf{0.408} \\ \bottomrule
    \end{tabular}
}
\caption{Accuracy ($\uparrow$), accuracy drop ($\uparrow$), and prediction-feature correlation ($\downarrow$) on four classification tasks of GPT-3 (Ada, 2.7B), finetuned with and without explanations.}
\vspace{-0.1in}
\label{tab: ada results}
\end{table*}

%% file: tables/babbage_table.tex
\begin{table*}[t]
\centering
\resizebox{\textwidth}{!}{%
\begin{tabular}{llrrrrrrrr}
    \toprule
     &  & \multicolumn{2}{c}{ComVE} & \multicolumn{2}{c}{CREAK} & \multicolumn{2}{c}{e-SNLI} & \multicolumn{2}{c}{SBIC} \\ \cline{3-10} 
     &  & Standard & Explain & Standard & Explain & Standard & Explain & Standard & Explain \\ \hline
    \multirow{6}{*}{
    \vspace{-2cm} 
    \begin{tabular}[c]{@{}l@{}}Accuracy\\ ($\delta_{acc}$)\end{tabular}
    } 
    & No Cue & \textbf{87.4} & 74.0 & \textbf{76.8} & 68.6 & \textbf{90.4} & 86.0 & 78.0 & \textbf{78.6} \\ \cline{2-10} 
& Sentence Length & \begin{tabular}[c]{@{}r@{}}50.4\\ \small (-37.0)\end{tabular} & \textbf{\begin{tabular}[c]{@{}r@{}}59.0\\ \small (-15.0)\end{tabular}} & \begin{tabular}[c]{@{}r@{}}52.8\\ \small (-24.0)\end{tabular} & \textbf{\begin{tabular}[c]{@{}r@{}}59.4\\ \small (-9.2)\end{tabular}} & \begin{tabular}[c]{@{}r@{}}\textbf{62.2}\\ \small (-28.2)\end{tabular} & \begin{tabular}[c]{@{}r@{}}60.6\\ \small \textbf{(-25.4)}\end{tabular} & \begin{tabular}[c]{@{}r@{}}51.2\\ \small (-26.8)\end{tabular} & \textbf{\begin{tabular}[c]{@{}r@{}}52.0\\ \small (-26.6)\end{tabular}} \\ \cline{2-10}
& Present Tense & \begin{tabular}[c]{@{}r@{}}54.2\\ \small (-33.2)\end{tabular} & \textbf{\begin{tabular}[c]{@{}r@{}}69.6\\ \small (-4.4)\end{tabular}} & \begin{tabular}[c]{@{}r@{}}55.8\\ \small (-21.0)\end{tabular} & \textbf{\begin{tabular}[c]{@{}r@{}}61.6\\ \small (-7.0)\end{tabular}} & \begin{tabular}[c]{@{}r@{}}75.0\\ \small (-15.4)\end{tabular} & \textbf{\begin{tabular}[c]{@{}r@{}}76.4\\ \small (-9.6)\end{tabular}} & \begin{tabular}[c]{@{}r@{}}73.6\\ \small (-4.4)\end{tabular} & \textbf{\begin{tabular}[c]{@{}r@{}}75.6\\ \small (-3.0)\end{tabular}} \\ \cline{2-10}
& Embedding Cluster & \begin{tabular}[c]{@{}r@{}}50.8\\ \small (-36.6)\end{tabular} & \textbf{\begin{tabular}[c]{@{}r@{}}55.6\\ \small (-18.4)\end{tabular}} & \begin{tabular}[c]{@{}r@{}}51.8\\ \small (-25.0)\end{tabular} & \textbf{\begin{tabular}[c]{@{}r@{}}55.4\\ \small (-13.2)\end{tabular}} & \begin{tabular}[c]{@{}r@{}}63.2\\ \small (-27.2)\end{tabular} & \textbf{\begin{tabular}[c]{@{}r@{}}69.0\\ \small (-17.0)\end{tabular}} & \begin{tabular}[c]{@{}r@{}}54.8\\ \small (-23.2)\end{tabular} & \textbf{\begin{tabular}[c]{@{}r@{}}56.6\\ \small (-22.0)\end{tabular}} \\ \cline{2-10}
& Plural Noun & \begin{tabular}[c]{@{}r@{}}52.8\\ \small (-34.6)\end{tabular} & \textbf{\begin{tabular}[c]{@{}r@{}}64.8\\ \small (-9.2)\end{tabular}} & \begin{tabular}[c]{@{}r@{}}54.4\\ \small (-22.4)\end{tabular} & \textbf{\begin{tabular}[c]{@{}r@{}}62.6\\ \small (-6.0)\end{tabular}} & \begin{tabular}[c]{@{}r@{}}59.8\\ \small (-30.6)\end{tabular} & \textbf{\begin{tabular}[c]{@{}r@{}}63.6\\ \small (-22.4)\end{tabular}} & \begin{tabular}[c]{@{}r@{}}75.8\\ \small (-2.2)\end{tabular} & \textbf{\begin{tabular}[c]{@{}r@{}}78.2\\ \small (-0.4)\end{tabular}} \\ \cline{2-10}
& Average & \begin{tabular}[c]{@{}r@{}}52.1\\ \small (-35.4)\end{tabular} & \textbf{\begin{tabular}[c]{@{}r@{}}62.3\\ \small (-11.8)\end{tabular}} & \begin{tabular}[c]{@{}r@{}}53.7\\ \small (-23.1)\end{tabular} & \textbf{\begin{tabular}[c]{@{}r@{}}59.8\\ \small (-8.8)\end{tabular}} & \begin{tabular}[c]{@{}r@{}}65.1\\ \small (-25.4)\end{tabular} & \textbf{\begin{tabular}[c]{@{}r@{}}67.4\\ \small (-18.6)\end{tabular}} & \begin{tabular}[c]{@{}r@{}}63.9\\ \small (-14.2)\end{tabular} & \textbf{\begin{tabular}[c]{@{}r@{}}65.6\\ \small (-13.0)\end{tabular}} \\ \hline
\multirow{5}{*}{\begin{tabular}[c]{@{}l@{}}Correlation between \\ Model's Prediction\\ and Spurious Feature\end{tabular}} & Sentence Length & 0.821 & \textbf{0.524} & 0.894 & \textbf{0.659} & 0.633 & \textbf{0.582} & 0.753 & \textbf{0.735} \\
& Present Tense & 0.791 & \textbf{0.528} & 0.704 & \textbf{0.465} & 0.439 & \textbf{0.341} & 0.417 & \textbf{0.269} \\
& Embedding Cluster & 0.815 & \textbf{0.675} & 0.735 & \textbf{0.665} & 0.761 & \textbf{0.484} & 0.570 & \textbf{0.551} \\
& Plural Noun & 0.838 & \textbf{0.494} & 0.714 & \textbf{0.373} & 0.721 & \textbf{0.579} & 0.220 & \textbf{0.191} \\
& Average & 0.816 & \textbf{0.555} & 0.762 & \textbf{0.541} & 0.639 & \textbf{0.496} & 0.490 & \textbf{0.437}
     \\ \bottomrule
    \end{tabular}
}
\caption{Accuracy ($\uparrow$), accuracy drop ($\uparrow$), and prediction-feature correlation ($\downarrow$) on four classification tasks of GPT-3 (Babbage), finetuned with and without explanations.}
\vspace{-0.1in}
\label{tab: babbage results}
\end{table*}

%% file: tables/curie_table.tex
\begin{table*}[t]
\centering
\resizebox{\textwidth}{!}{%
\begin{tabular}{llrrrrrrrr}
    \toprule
     &  & \multicolumn{2}{c}{ComVE} & \multicolumn{2}{c}{CREAK} & \multicolumn{2}{c}{e-SNLI} & \multicolumn{2}{c}{SBIC} \\ \cline{3-10} 
     &  & Standard & Explain & Standard & Explain & Standard & Explain & Standard & Explain \\ \hline
    \multirow{6}{*}{
    \vspace{-2cm} 
    \begin{tabular}[c]{@{}l@{}}Accuracy\\ ($\delta_{acc}$)\end{tabular}
    } & No Cue & \textbf{92.2} & 84.2 & \textbf{83.2} & 76.0 & \textbf{91.4} & 88.2 & \textbf{78.8} & 76.8 \\ \cline{2-10} 
     & Sentence Length & \begin{tabular}[c]{@{}r@{}}56.2\\ \small (-36.0)\end{tabular} & \textbf{\begin{tabular}[c]{@{}r@{}}73.6\\ \small (-10.6)\end{tabular}} & \begin{tabular}[c]{@{}r@{}}54.8\\ \small (-28.4)\end{tabular} & \textbf{\begin{tabular}[c]{@{}r@{}}70.4\\ \small (-5.6)\end{tabular}} & \textbf{\begin{tabular}[c]{@{}r@{}}78.2\\ \small (-13.2)\end{tabular}} & \begin{tabular}[c]{@{}r@{}}73.0\\ \small (-15.2)\end{tabular} & \begin{tabular}[c]{@{}r@{}}\textbf{53.0}\\ \small (-25.8)\end{tabular} & \begin{tabular}[c]{@{}r@{}}52.0\\ \small \textbf{(-24.8)}\end{tabular} \\ \cline{2-10} 
     & Present Tense & \begin{tabular}[c]{@{}r@{}}62.8\\ \small (-29.4)\end{tabular} & \textbf{\begin{tabular}[c]{@{}r@{}}82.2\\ \small (-2.0)\end{tabular}} & \begin{tabular}[c]{@{}r@{}}62.8\\ \small (-20.4)\end{tabular} & \textbf{\begin{tabular}[c]{@{}r@{}}69.8\\ \small (-6.2)\end{tabular}} & \begin{tabular}[c]{@{}r@{}}79.2\\ \small (-12.2)\end{tabular} & \textbf{\begin{tabular}[c]{@{}r@{}}88.6\\ \small (-0.4)\end{tabular}} & \textbf{\begin{tabular}[c]{@{}r@{}}77.2\\ \small (-1.6)\end{tabular}} & \begin{tabular}[c]{@{}r@{}}76.4\\ \small (-1.84)\end{tabular} \\ \cline{2-10} 
     & Embedding Cluster & \begin{tabular}[c]{@{}r@{}}70.0\\ \small (-22.2)\end{tabular} & \textbf{\begin{tabular}[c]{@{}r@{}}70.6\\ \small (-13.6)\end{tabular}} & \begin{tabular}[c]{@{}r@{}}58.2\\ \small (-25.0)\end{tabular} & \textbf{\begin{tabular}[c]{@{}r@{}}60.6\\ \small (-15.4)\end{tabular}} & \begin{tabular}[c]{@{}r@{}}63.4\\ \small (-28.0)\end{tabular} & \textbf{\begin{tabular}[c]{@{}r@{}}82.6\\ \small (-5.6)\end{tabular}} & \begin{tabular}[c]{@{}r@{}}57.8\\ \small (-21.0)\end{tabular} & \begin{tabular}[c]{@{}r@{}}\textbf{58.4}\\ \small \textbf{(-18.4)}\end{tabular} \\ \cline{2-10} 
     & Plural Noun & \begin{tabular}[c]{@{}r@{}}65.0\\ \small (-27.2)\end{tabular} & \textbf{\begin{tabular}[c]{@{}r@{}}82.2\\ \small (-2.0)\end{tabular}} & \begin{tabular}[c]{@{}r@{}}60.0\\ \small (-23.2)\end{tabular} & \textbf{\begin{tabular}[c]{@{}r@{}}73.0\\ \small (-3.0)\end{tabular}} & \begin{tabular}[c]{@{}r@{}}59.0\\ \small (-32.4)\end{tabular} & \textbf{\begin{tabular}[c]{@{}r@{}}78.4\\ \small (-9.8)\end{tabular}} & \begin{tabular}[c]{@{}r@{}}76.2\\ \small (-2.6)\end{tabular} & \textbf{\begin{tabular}[c]{@{}r@{}}77.0\\ \small (0.2)\end{tabular}} \\ \cline{2-10} 
     & Average & \begin{tabular}[c]{@{}r@{}}63.5\\ \small (-28.7)\end{tabular} & \begin{tabular}[c]{@{}r@{}}\textbf{77.2}\\ \small \textbf{(-7.1)}\end{tabular} & \begin{tabular}[c]{@{}r@{}}59.0\\ \small (-24.3)\end{tabular} & \begin{tabular}[c]{@{}r@{}}\textbf{68.5}\\ \small \textbf{(-7.6)}\end{tabular} & \begin{tabular}[c]{@{}r@{}}70.0\\ \small (-21.5)\end{tabular} & \textbf{\begin{tabular}[c]{@{}r@{}}80.7\\ \small (-7.6)\end{tabular}} & \begin{tabular}[c]{@{}r@{}}66.1\\ \small (-12.8)\end{tabular} & \begin{tabular}[c]{@{}r@{}}\textbf{66.0}\\ \small \textbf{(-10.9)}\end{tabular} \\ \hline
    \multirow{5}{*}{\begin{tabular}[c]{@{}l@{}}Correlation between \\ Model's Prediction\\ and Spurious Feature\end{tabular}} & Sentence Length & 0.736 & \textbf{0.347} & 0.872 & \textbf{0.413} &\textbf{ 0.305} & 0.333 & \textbf{0.684} & 0.701 \\
     & Present Tense & 0.756 & \textbf{0.244} & 0.589 & \textbf{0.402} & 0.364 & \textbf{0.075} & 0.244 & \textbf{0.231} \\ 
     & Embedding Cluster & 0.444 & \textbf{0.426} & 0.678 & \textbf{0.533} & 0.738 & \textbf{0.226} & \textbf{0.386} & 0.418 \\ 
     & Plural Noun & 0.594 & \textbf{0.267} & 0.570 & \textbf{0.208} & 0.777 & \textbf{0.278} & 0.183 & \textbf{0.114} \\ 
     & Average & 0.633 & \textbf{0.321} & 0.677 & \textbf{0.389} & 0.546 & \textbf{0.228} & 0.399 & \textbf{0.366} \\ \bottomrule
    \end{tabular}
}
\caption{Accuracy ($\uparrow$), accuracy drop ($\uparrow$), and prediction-feature correlation ($\downarrow$) on four classification tasks of GPT-3 (Curie), finetuned with and without explanations.}
\vspace{-0.1in}
\label{tab: curie results}
\end{table*}

%% file: tables/t5_table.tex
\begin{table*}[t]
\centering
\resizebox{\textwidth}{!}{%
\begin{tabular}{llrrrrrrrr}
    \toprule
     &  & \multicolumn{2}{c}{ComVE} & \multicolumn{2}{c}{CREAK} & \multicolumn{2}{c}{e-SNLI} & \multicolumn{2}{c}{SBIC} \\ \cline{3-10} 
     &  & Standard & Explain & Standard & Explain & Standard & Explain & Standard & Explain \\ \hline
    \multirow{6}{*}{
    \vspace{-2cm} 
    \begin{tabular}[c]{@{}l@{}}Accuracy\\ ($\delta_{acc}$)\end{tabular}
    } & No Cue & \textbf{76.4} & 49.8 & \textbf{55.2} & 41.4 & \textbf{86.6} & 55.6 & \textbf{69.4} & 65.0 \\ \cline{2-10} 
     & Sentence Length & \begin{tabular}[c]{@{}r@{}}\textbf{53.6}\\ \small (-22.8)\end{tabular} & \begin{tabular}[c]{@{}r@{}}51.2\\ \textbf{\small (1.4)}\end{tabular} & \begin{tabular}[c]{@{}r@{}}\textbf{52.6}\\ \small (-2.6)\end{tabular} & \begin{tabular}[c]{@{}r@{}}45.6\\ \textbf{\small (4.2)}\end{tabular} & \begin{tabular}[c]{@{}r@{}}\textbf{64.0}\\ \small (-22.6)\end{tabular} & \begin{tabular}[c]{@{}r@{}}51.6\\ \textbf{\small (-4.0)}\end{tabular} & \begin{tabular}[c]{@{}r@{}}\textbf{56.0}\\ \small (-13.4)\end{tabular} & \begin{tabular}[c]{@{}r@{}}53.4\\ \textbf{\small (-11.6)}\end{tabular} \\ \cline{2-10} 
     & Present Tense & \begin{tabular}[c]{@{}r@{}}\textbf{61.6}\\ \small (-14.8)\end{tabular} & \begin{tabular}[c]{@{}r@{}}51.2\\ \textbf{\small (1.4)}\end{tabular} & \begin{tabular}[c]{@{}r@{}}\textbf{50.0}\\ \small (-5.2)\end{tabular} & \begin{tabular}[c]{@{}r@{}}41.8\\ \textbf{\small (0.4)}\end{tabular} & \textbf{\begin{tabular}[c]{@{}r@{}}79.4\\ \small (-7.2)\end{tabular}} & \begin{tabular}[c]{@{}r@{}}42.6\\ \small (-13.0)\end{tabular} & \textbf{\begin{tabular}[c]{@{}r@{}}70.6\\ \small (1.2)\end{tabular}} & \begin{tabular}[c]{@{}r@{}}63.6\\ \small (-1.4)\end{tabular} \\ \cline{2-10} 
     & Embedding Cluster & \begin{tabular}[c]{@{}r@{}}\textbf{59.4}\\ \small (-17.0)\end{tabular} & \begin{tabular}[c]{@{}r@{}}44.6\\ \textbf{\small (-5.2)}\end{tabular} & \begin{tabular}[c]{@{}r@{}}\textbf{49.4}\\ \small (-5.8)\end{tabular} & \begin{tabular}[c]{@{}r@{}}38.4\\ \textbf{\small (-3.0)}\end{tabular} & \begin{tabular}[c]{@{}r@{}}\textbf{69.8}\\ \small (-16.8)\end{tabular} & \begin{tabular}[c]{@{}r@{}}42.6\\ \textbf{\small (-13.0)}\end{tabular} & \textbf{\begin{tabular}[c]{@{}r@{}}71.8\\ \small (2.4)\end{tabular}} & \begin{tabular}[c]{@{}r@{}}64.0\\ \small (-1.0)\end{tabular} \\ \cline{2-10} 
     & Plural Noun & \begin{tabular}[c]{@{}r@{}}\textbf{73.8}\\ \small (-2.6)\end{tabular} & \begin{tabular}[c]{@{}r@{}}53.4\\ \textbf{\small (3.6)}\end{tabular} & \begin{tabular}[c]{@{}r@{}}\textbf{50.8}\\ \small (-4.4)\end{tabular} & \begin{tabular}[c]{@{}r@{}}40.6\\ \textbf{\small (-0.8)}\end{tabular} & \begin{tabular}[c]{@{}r@{}}\textbf{59.4}\\ \small (-27.2)\end{tabular} & \begin{tabular}[c]{@{}r@{}}43.8\\ \textbf{\small (-11.8)}\end{tabular} & \begin{tabular}[c]{@{}r@{}}\textbf{69.4}\\ \small (0.0)\end{tabular} & \begin{tabular}[c]{@{}r@{}}66.4\\ \textbf{\small (1.4)}\end{tabular} \\ \cline{2-10} 
     & Average & \begin{tabular}[c]{@{}r@{}}\textbf{62.1}\\ \small (-14.3)\end{tabular} & \begin{tabular}[c]{@{}r@{}}50.1\\ \textbf{\small (0.3)}\end{tabular} & \begin{tabular}[c]{@{}r@{}}\textbf{50.7}\\ \small (-4.5)\end{tabular} & \begin{tabular}[c]{@{}r@{}}41.6\\ \textbf{\small (0.2)}\end{tabular} & \begin{tabular}[c]{@{}r@{}}\textbf{68.2}\\ \small (-18.5)\end{tabular} & \begin{tabular}[c]{@{}r@{}}45.2\\ \textbf{\small (-10.5)}\end{tabular} & \textbf{\begin{tabular}[c]{@{}r@{}}67.0\\ \small (-2.5)\end{tabular}} & \begin{tabular}[c]{@{}r@{}}61.9\\ \small (-3.2)\end{tabular} \\ \hline
    \multirow{5}{*}{\begin{tabular}[c]{@{}l@{}}Prediction-\\Feature \\ Correlation\end{tabular}} & Sentence Length & 0.641 & \textbf{0.402} & 0.699 & \textbf{0.115} & 0.524 & \textbf{0.384} & \textbf{0.222} & 0.376 \\ 
     & Present Tense & 0.653 & \textbf{0.166} & 0.575 & \textbf{0.513} & 0.281 & \textbf{0.231} & \textbf{0.217} & 0.319 \\
     & Embedding Cluster & 0.645 & \textbf{0.463} & 0.694 & \textbf{0.456} & 0.494 & \textbf{0.169} & 0.504 & \textbf{0.473} \\
     & Plural Noun & 0.343 & \textbf{0.176} & 0.481 & \textbf{0.269} & 0.722 & \textbf{0.207} & \textbf{0.107} & 0.205 \\
     & Average & 0.571 & \textbf{0.302} & 0.612 & \textbf{0.338} & 0.505 & \textbf{0.248} & \textbf{0.263} & 0.343 \\ \bottomrule
    \end{tabular}
}
\caption{Accuracy ($\uparrow$), accuracy drop ($\uparrow$), and prediction-feature correlation ($\downarrow$) on four classification tasks of T5-base, finetuned with and without explanations.}
\label{tab: t5-base}
\end{table*}

%% file: tables/bart_table.tex
\begin{table*}[t]
\centering
\resizebox{\textwidth}{!}{%
\begin{tabular}{llrrrrrrrr}
    \toprule
     &  & \multicolumn{2}{c}{ComVE} & \multicolumn{2}{c}{CREAK} & \multicolumn{2}{c}{e-SNLI} & \multicolumn{2}{c}{SBIC} \\ \cline{3-10} 
     &  & Standard & Explain & Standard & Explain & Standard & Explain & Standard & Explain \\ \hline
    \multirow{6}{*}{
    \vspace{-2cm} 
    \begin{tabular}[c]{@{}l@{}}Accuracy\\ ($\delta_{acc}$)\end{tabular}
    } & No Cue & \textbf{53.2} & 48.0 & \textbf{59.0} & 46.0 & \textbf{85.6} & 46.2 & \textbf{76.8} & 51.0 \\ \cline{2-10} 
     & Sentence Length & \begin{tabular}[c]{@{}r@{}}42.8\\ \small (-10.4)\end{tabular} & \textbf{\begin{tabular}[c]{@{}r@{}}43.6\\ \small (-4.4)\end{tabular}} & \begin{tabular}[c]{@{}r@{}}\textbf{54.6}\\ \small (-4.4)\end{tabular} & \begin{tabular}[c]{@{}r@{}}48.4\\ \small \textbf{(2.4)}\end{tabular} & \begin{tabular}[c]{@{}r@{}}\textbf{54.4}\\ \small (-31.2)\end{tabular} & \begin{tabular}[c]{@{}r@{}}43.4\\ \small \textbf{(-2.8)}\end{tabular} & \begin{tabular}[c]{@{}r@{}}\textbf{50.8}\\ \small (-26.0)\end{tabular} & \begin{tabular}[c]{@{}r@{}}50.2\\ \small \textbf{(-0.8)}\end{tabular} \\ \cline{2-10} 
     & Present Tense & \begin{tabular}[c]{@{}r@{}}\textbf{55.2}\\ \small (2.0)\end{tabular} & \begin{tabular}[c]{@{}r@{}}54.0\\ \small \textbf{(6.0)}\end{tabular} & \begin{tabular}[c]{@{}r@{}}\textbf{53.2}\\ \small (-5.8)\end{tabular} & \begin{tabular}[c]{@{}r@{}}48.0\\ \small \textbf{(2.0)}\end{tabular} & \begin{tabular}[c]{@{}r@{}}\textbf{58.8}\\ \small (-26.8)\end{tabular} & \begin{tabular}[c]{@{}r@{}}44.0\\ \small \textbf{(-2.2)}\end{tabular} & \begin{tabular}[c]{@{}r@{}}\textbf{70.8}\\ \small (-6.0)\end{tabular} & \begin{tabular}[c]{@{}r@{}}63.0\\ \small \textbf{(12.0)}\end{tabular} \\ \cline{2-10} 
     & Embedding Cluster & \begin{tabular}[c]{@{}r@{}}\textbf{48.0}\\ \small (-5.2)\end{tabular} & \begin{tabular}[c]{@{}r@{}}47.2\\ \small \textbf{(-0.8)}\end{tabular} & \begin{tabular}[c]{@{}r@{}}\textbf{49.6}\\ \small (-9.4)\end{tabular} & \begin{tabular}[c]{@{}r@{}}44.0\\ \small \textbf{(-2.0)}\end{tabular} & \begin{tabular}[c]{@{}r@{}}\textbf{54.0}\\ \small (-31.6)\end{tabular} & \begin{tabular}[c]{@{}r@{}}40.6\\ \small \textbf{(-5.6)}\end{tabular} & \begin{tabular}[c]{@{}r@{}}\textbf{68.6}\\ \small (-8.2)\end{tabular} & \begin{tabular}[c]{@{}r@{}}60.0\\ \small \textbf{(9.0)}\end{tabular} \\ \cline{2-10} 
     & Plural Noun & \begin{tabular}[c]{@{}r@{}}\textbf{54.0}\\ \small (0.8)\end{tabular} & \begin{tabular}[c]{@{}r@{}}51.2\\ \small \textbf{(3.2)}\end{tabular} & \begin{tabular}[c]{@{}r@{}}\textbf{53.2}\\ \small (-5.8)\end{tabular} & \begin{tabular}[c]{@{}r@{}}46.8\\ \small \textbf{(0.8)}\end{tabular} & \begin{tabular}[c]{@{}r@{}}\textbf{52.8}\\ \small (-32.8)\end{tabular} & \begin{tabular}[c]{@{}r@{}}48.4\\ \small \textbf{(2.2)}\end{tabular} & \begin{tabular}[c]{@{}r@{}}\textbf{65.2}\\ \small (-11.6)\end{tabular} & \begin{tabular}[c]{@{}r@{}}53.8\\ \small \textbf{(2.8)}\end{tabular} \\ \cline{2-10} 
     & Average & \begin{tabular}[c]{@{}r@{}}\textbf{50.0}\\ \small (-3.2)\end{tabular} & \begin{tabular}[c]{@{}r@{}}49.0\\ \small \textbf{(1.0)}\end{tabular} & \begin{tabular}[c]{@{}r@{}}\textbf{52.7}\\ \small (-6.4)\end{tabular} & \begin{tabular}[c]{@{}r@{}}46.8\\ \small \textbf{(0.8)}\end{tabular} & \begin{tabular}[c]{@{}r@{}}\textbf{55.0}\\ \small (-30.6)\end{tabular} & \begin{tabular}[c]{@{}r@{}}44.1\\ \small \textbf{(-2.1)}\end{tabular} & \begin{tabular}[c]{@{}r@{}}\textbf{63.9}\\ \small (-13.0)\end{tabular} & \begin{tabular}[c]{@{}r@{}}56.8\\ \small \textbf{(5.8)}\end{tabular} \\ \hline
    \multirow{5}{*}{\begin{tabular}[c]{@{}l@{}}Prediction-\\Feature \\ Correlation\end{tabular}} & Sentence Length & 0.667 & \textbf{0.638} & 0.762 & \textbf{0.629} & \textbf{0.724} & 0.745 & \textbf{0.288} & 0.706 \\ 
     & Present Tense & 0.881 & \textbf{0.744} & 0.603 & \textbf{0.454} & 0.702 & \textbf{0.159} & \textbf{0.241} & 0.314 \\
     & Embedding Cluster & 0.817 & \textbf{0.792} & 0.801 & \textbf{0.700} & 0.854 & \textbf{0.301} & 0.555 & \textbf{0.395} \\
     & Plural Noun & 0.823 & \textbf{0.230} & 0.607 & \textbf{0.491} & 0.884 & \textbf{0.439} & 0.287 & \textbf{0.210} \\ 
     & Average & 0.797 & \textbf{0.601} & 0.693 & \textbf{0.569} & 0.791 & \textbf{0.411} & \textbf{0.343} & 0.406 \\ \bottomrule
    \end{tabular}
}
\caption{Accuracy ($\uparrow$), accuracy drop ($\uparrow$), and prediction-feature correlation ($\downarrow$) on four classification tasks of BART-base, finetuned with and without explanations.}
\label{tab: bart-base}
\end{table*}

%% file: tables/opt_table.tex
\begin{table*}[t]
    \centering
    \resizebox{\textwidth}{!}{%
    \begin{tabular}{llrrrrrrrr}
    \toprule
         &  & \multicolumn{2}{c}{ComVE} & \multicolumn{2}{c}{CREAK} & \multicolumn{2}{c}{e-SNLI} & \multicolumn{2}{c}{SBIC} \\ \cline{3-10} 
         &  & Standard & Explain & Standard & Explain & Standard & Explain & Standard & Explain \\ \hline
        \multirow{6}{*}{
        \vspace{-2cm} 
        \begin{tabular}[c]{@{}l@{}}Accuracy\\ ($\delta_{acc}$)\end{tabular}
        } 
        & No Cue & 78.4 & \textbf{82.6} & \textbf{75.2} & 66.0 & \textbf{86.8} & 72.7 & \textbf{76.0} & 73.6 \\ \cline{2-10} 
     & Sentence Length & \begin{tabular}[c]{@{}r@{}}50.6\\ \small (-27.8)\end{tabular} & \textbf{\begin{tabular}[c]{@{}r@{}}63.4\\ \small (-19.2)\end{tabular}} & \begin{tabular}[c]{@{}r@{}}56.4\\ \small (-18.8)\end{tabular} & \textbf{\begin{tabular}[c]{@{}r@{}}62.0\\ \small (-4.0)\end{tabular}} & \textbf{\begin{tabular}[c]{@{}r@{}}73.0\\ \small (-13.8)\end{tabular}} & \begin{tabular}[c]{@{}r@{}}56.4\\ \small (-16.3)\end{tabular} & \begin{tabular}[c]{@{}r@{}}54.0\\ \small (-22.0)\end{tabular} & \textbf{\begin{tabular}[c]{@{}r@{}}56.2\\ \small (-17.4)\end{tabular}} \\ \cline{2-10} 
     & Present Tense & \begin{tabular}[c]{@{}r@{}}62.6\\ \small (-15.8)\end{tabular} & \textbf{\begin{tabular}[c]{@{}r@{}}70.2\\ \small (-12.4)\end{tabular}} & \begin{tabular}[c]{@{}r@{}}\textbf{66.8}\\ \small (-8.4)\end{tabular} & \begin{tabular}[c]{@{}r@{}}61.4\\ \small \textbf{(-4.6)}\end{tabular} & \begin{tabular}[c]{@{}r@{}}\textbf{72.4}\\ \small (-14.4)\end{tabular} & \begin{tabular}[c]{@{}r@{}}67.8\\ \small \textbf{(-4.9)}\end{tabular} & \textbf{\begin{tabular}[c]{@{}r@{}}72.0\\ \small (-4.0)\end{tabular}} & \begin{tabular}[c]{@{}r@{}}66.2\\ \small (-7.4)\end{tabular} \\ \cline{2-10}
     & Embedding Cluster & \begin{tabular}[c]{@{}r@{}}53.4\\ \small (-25.0)\end{tabular} & \textbf{\begin{tabular}[c]{@{}r@{}}58.6\\ \small (-24.0)\end{tabular}} & \begin{tabular}[c]{@{}r@{}}53.6\\ \small (-21.6)\end{tabular} & \textbf{\begin{tabular}[c]{@{}r@{}}55.2\\ \small (-10.8)\end{tabular}} & \begin{tabular}[c]{@{}r@{}}55.4\\ \small (-31.4)\end{tabular} & \textbf{\begin{tabular}[c]{@{}r@{}}56.0\\ \small (-16.7)\end{tabular}} & \begin{tabular}[c]{@{}r@{}}63.6\\ \small (-12.4)\end{tabular} & \textbf{\begin{tabular}[c]{@{}r@{}}64.8\\ \small (-8.8)\end{tabular}} \\ \cline{2-10}
     & Plural Noun & \begin{tabular}[c]{@{}r@{}}65.6\\ \small \textbf{(-12.8)}\end{tabular} & \begin{tabular}[c]{@{}r@{}}\textbf{67.2}\\ \small (-15.4)\end{tabular} & \begin{tabular}[c]{@{}r@{}}60.0\\ \small (-15.2)\end{tabular} & \textbf{\begin{tabular}[c]{@{}r@{}}61.6\\ \small (-4.4)\end{tabular}} & \begin{tabular}[c]{@{}r@{}}54.8\\ \small (-32.0)\end{tabular} & \textbf{\begin{tabular}[c]{@{}r@{}}58.2\\ \small (-14.5)\end{tabular}} & \textbf{\begin{tabular}[c]{@{}r@{}}72.6\\ \small (-3.4)\end{tabular}} & \begin{tabular}[c]{@{}r@{}}69.8\\ \small (-3.8)\end{tabular} \\ \cline{2-10}
     & Average & \begin{tabular}[c]{@{}r@{}}58.1\\ \small (-20.4)\end{tabular} & \textbf{\begin{tabular}[c]{@{}r@{}}64.9\\ \small (-17.8)\end{tabular}} & \begin{tabular}[c]{@{}r@{}}59.2\\ \small (-16.0)\end{tabular} & \textbf{\begin{tabular}[c]{@{}r@{}}60.1\\ \small (-6.0)\end{tabular}} & \begin{tabular}[c]{@{}r@{}}\textbf{63.9}\\ \small (-22.9)\end{tabular} & \begin{tabular}[c]{@{}r@{}}59.6\\ \small \textbf{(-13.1)}\end{tabular} & \begin{tabular}[c]{@{}r@{}}\textbf{65.6}\\ \small (-10.5)\end{tabular} & \begin{tabular}[c]{@{}r@{}}64.3\\ \small \textbf{(-9.3)}\end{tabular} \\ \hline
     \multirow{5}{*}{\begin{tabular}[c]{@{}l@{}}Correlation between \\ Model's Prediction\\ and Spurious Feature\end{tabular}} 
     & Sentence Length & 0.376 & \textbf{0.221} & 0.784 & \textbf{0.286} & 0.237 & \textbf{0.102} & 0.093 & \textbf{0.008} \\
     & Present Tense & 0.686 & \textbf{0.282} & 0.499 & \textbf{0.437} & 0.419 & \textbf{0.331} & 0.224 & \textbf{0.187} \\
     & Embedding Cluster & 0.693 & \textbf{0.571} & 0.727 & \textbf{0.529} & 0.852 & \textbf{0.555} & 0.397 & \textbf{0.319} \\
     & Plural Noun & 0.641 & \textbf{0.183} & 0.385 & \textbf{0.218} & 0.762 & \textbf{0.463} & \textbf{0.114} & 0.129 \\
    & Average & 0.599 & \textbf{0.314} & 0.599 & \textbf{0.368} & 0.568 & \textbf{0.363} & 0.207 & \textbf{0.161} \\ \bottomrule
    \end{tabular}
    }

    \caption{Accuracy ($\uparrow$), accuracy drop ($\uparrow$), and prediction-feature correlation ($\downarrow$) on four classification tasks of OPT (1.3b), finetuned with and without explanations.}
\vspace{-0.1in}
\label{tab: opt results}

    \end{table*}

%% file: tables/fewshot_table.tex
\begin{table*}[]
\centering
\resizebox{\textwidth}{!}{%
\begin{tabular}{llrrr}
\toprule
 &  & \multicolumn{1}{p{3cm}}{Standard-finetuned model} & \multicolumn{1}{p{4cm}}{Standard prompting on standard-finetuned model} & \multicolumn{1}{p{5cm}}{Explanation-based prompting on standard-finetuned model} \\ \hline
\multirow{6}{*}{\vspace{-2cm} 
    \begin{tabular}[c]{@{}l@{}}Accuracy\\ ($\delta_{acc}$)\end{tabular}} & No cue & 88.0 & N/A & N/A \\ \cline{2-5}
& Sentence Length & \begin{tabular}[c]{@{}r@{}}69.8\\ \small (-18.2)\end{tabular} & \textbf{\begin{tabular}[c]{@{}r@{}}78.4\\ \small (-9.6)\end{tabular}} & \begin{tabular}[c]{@{}r@{}}72.4\\ \small (-15.6)\end{tabular} \\ \cline{2-5} 
& Present Tense & \begin{tabular}[c]{@{}r@{}}76.0\\ \small (-12.0)\end{tabular} & \textbf{\begin{tabular}[c]{@{}r@{}}86.0\\ \small (-2.0)\end{tabular}} & \begin{tabular}[c]{@{}r@{}}83.6\\ \small (-4.4)\end{tabular} \\ \cline{2-5}
& Embedding Cluster & \begin{tabular}[c]{@{}r@{}}70.6\\ \small (-17.4)\end{tabular} & \textbf{\begin{tabular}[c]{@{}r@{}}71.2\\ \small (-16.8)\end{tabular}} & \begin{tabular}[c]{@{}r@{}}62.8\\ \small (-25.2)\end{tabular} \\ \cline{2-5}
& Plural Noun & \begin{tabular}[c]{@{}r@{}}69.0\\ \small (-19.0)\end{tabular} & \textbf{\begin{tabular}[c]{@{}r@{}}83.6\\ \small (-4.4)\end{tabular}} & \begin{tabular}[c]{@{}r@{}}78.6\\ \small (-9.4)\end{tabular} \\ \cline{2-5}
& Average & \begin{tabular}[c]{@{}r@{}}71.4\\ \small (-16.7)\end{tabular} & \begin{tabular}[c]{@{}r@{}}\textbf{79.8}\\ \small \textbf{(-8.2)}\end{tabular} & \begin{tabular}[c]{@{}r@{}}74.4\\ \small (-13.7)\end{tabular} \\ \hline
\multirow{5}{*}{\begin{tabular}[c]{@{}l@{}}Prediction-\\Feature \\ Correlation\end{tabular}} & Sentence Length & 0.467 & \textbf{0.109} & 0.148 \\
 & Present Tense & 0.336 & \textbf{0.039} & 0.085 \\
 & Embedding Cluster & 0.595 & \textbf{0.532} & 0.691 \\
 & Plural Noun & 0.578 & \textbf{0.219} & 0.304 \\
 & Average & 0.494 & \textbf{0.225} & 0.307 \\
 \bottomrule
\end{tabular}}
\caption{Standard few-shot prompting vs. explanation-based few-shot prompting on standard-finetuned model with e-SNLI dataset. The accuracy difference $\delta_{acc}$ for the last two columns are based on the standard-finetuned model under the ``no cue'' setting.}
\label{tab: few-shot-over-finetuned experiments}
\end{table*}

%% file: tables/esnli_cue_strength_table.tex
\begin{table}[t]
\centering
\resizebox{\columnwidth}{!}{%
\begin{tabular}{llrrrr}
\toprule
 &  & \multicolumn{2}{c}{1k Examples} & \multicolumn{2}{c}{4k Examples} \\ \cline{3-6}
 &  & Standard & Explain & Standard & Explain \\ \hline
\multirow{6}{*}{
\vspace{-0.3cm} 
\begin{tabular}[c]{@{}l@{}}Accuracy\\ ($\delta_{acc}$)\end{tabular}
} & 0.2 & \textbf{91.4} & 86.8 & \textbf{94.4} & 89.8 \\
 & 0.6 & \textbf{85.8} & 82.8 & \textbf{90.8} & 90.6 \\
 & 0.8 & 84.2 & \textbf{87.0} & 88.4 & \textbf{89.8} \\
 & 0.9 & 81.0 & \textbf{86.8} & 83.2 & \textbf{91.0} \\
 & 1.0 & 61.4 & \textbf{79.8} & 58.8 & \textbf{87.6} \\
 & Average & 80.8 & \textbf{84.6} & 83.1 & \textbf{89.8} \\ \hline
\multirow{6}{*}{\begin{tabular}[c]{@{}l@{}}Prediction-\\Feature \\ Correlation\end{tabular}} & 0.2 & \textbf{0.044} & 0.097 & \textbf{0.024} & 0.094 \\
 & 0.6 & 0.211 & \textbf{0.147} & 0.160 & \textbf{0.117} \\
 & 0.8 & 0.268 & \textbf{0.113} & 0.231 & \textbf{0.158} \\
 & 0.9 & 0.367 & \textbf{0.130} & 0.336 & \textbf{0.141} \\
 & 1.0 & 0.769 & \textbf{0.239} & 0.84 & \textbf{0.233} \\
 & Average & 0.332 & \textbf{0.145} & 0.318 & \textbf{0.149} \\ \bottomrule
\end{tabular}
}
\caption{Accuracy ($\uparrow$) and prediction-feature correlation ($\downarrow$) of GPT-3 (Davinci) on e-SNLI, as the strength of the ``embedding cluster'' spurious correlation and the number of training examples varies.}
\label{tab: n-finetune-bias-strength}
\end{table}


%% file: tables/sbic_comve_finetune_n.tex
\begin{table}[]
\centering
\resizebox{\columnwidth}{!}{%
\begin{tabular}{lllrrrr}
\toprule
 &  &  & \multicolumn{2}{c}{ComVE} & \multicolumn{2}{c}{SBIC} \\\cline{4-7}
 &  &  & Standard & Explain & Standard & Explain \\ \hline
\multirow{4}{*}{\begin{tabular}[c]{@{}l@{}}Accuracy\\ ($\delta_{acc}$)\end{tabular}} & \multirow{2}{*}{\begin{tabular}[c]{@{}l@{}}Present \\Tense\end{tabular}} & n=1k & \textbf{93.6} & 89.4 & \textbf{78.6} & 77.4 \\ \cline{3-7}
 &  & n=4k & 81.6 & \textbf{94.8} & 65.4 & \textbf{77.0} \\ \cline{2-7}
 & \multirow{2}{*}{\begin{tabular}[c]{@{}l@{}}Sentence \\Length\end{tabular}} & n=1k & \textbf{91.4} & 89.4 & 50.4 & \textbf{53.4} \\ \cline{3-7}
 &  & n=4k & 83.2 & \textbf{89.0} & 50.4 & \textbf{53.4} \\ \hline
\multirow{4}{*}{\begin{tabular}[c]{@{}l@{}}Prediction-\\Feature \\ Correlation\end{tabular}} & \multirow{2}{*}{\begin{tabular}[c]{@{}l@{}}Present \\Tense\end{tabular}} & n=1k & 0.074 & \textbf{0.035} & 0.241 & \textbf{0.166} \\ \cline{3-7}
 &  & n=4k & 0.316 & \textbf{0.021} & 0.387 & \textbf{-0.001} \\ \cline{2-7}
 & \multirow{2}{*}{\begin{tabular}[c]{@{}l@{}}Sentence \\Length\end{tabular}} & n=1k & 0.134 & \textbf{0.108} & 0.732 & \textbf{0.166} \\ \cline{3-7}
 &  & n=4k & 0.245 & \textbf{0.109} & 0.770 & \textbf{0.670} \\ \bottomrule
\end{tabular}
}
\caption{Standard finetuning vs. explanation-based finetuning on selected settings after increasing number of examples.}
\label{tab: n-increase experiments}
\end{table}

%% file: tables/domain_sepcific_davinci_table.tex
\begin{table*}[t]
    \centering
    \resizebox{\textwidth}{!}{%
    \begin{tabular}{llrrrrrrrr}
    \toprule
     &  & \multicolumn{2}{c}{ComVE} & \multicolumn{2}{c}{CREAK} & \multicolumn{2}{c}{e-SNLI} & \multicolumn{2}{c}{SBIC} \\ \cline{3-10} 
     &  & Standard & Explain & Standard & Explain & Standard & Explain & Standard & Explain \\ \hline
    \multirow{2}{*}{
    \vspace{-0.5cm} 
    \begin{tabular}[c]{@{}l@{}}Accuracy\\ ($\delta_{acc}$)\end{tabular}
    } & No Cue & \textbf{97.0} & 95.6 & 84.2 & \textbf{85.0} & \textbf{91.6 }& 89.2 & \textbf{79.0} & 75.0 \\ \cline{2-10} 
     & Domain Specific & \textbf{\begin{tabular}[c]{@{}r@{}}93.6\\ \small (-3.4)\end{tabular}} & \begin{tabular}[c]{@{}r@{}}90.4\\ \small (-5.3)\end{tabular} & \textbf{\begin{tabular}[c]{@{}r@{}}80.5\\ \small  (-3.7)\end{tabular}} & \begin{tabular}[c]{@{}r@{}}79.0\\ \small (-6.0)\end{tabular} & \begin{tabular}[c]{@{}r@{}}55.8 \\ \small (-35.8)\end{tabular} & \textbf{\begin{tabular}[c]{@{}r@{}}86.6\\ \small (-2.6)\end{tabular}} & \begin{tabular}[c]{@{}r@{}}\textbf{42.6}\\ \small  \textbf{(-36.4)}\end{tabular} & \begin{tabular}[c]{@{}r@{}}\textbf{}38.3\\ \small (-36.7)\end{tabular} \\ \hline
    \begin{tabular}[c]{@{}l@{}}Prediction-\\Feature \\ Correlation\end{tabular} & Domain Specific & \textbf{0.055} & 0.097 &  0.112 & \textbf{-0.026} & 0.684 & \textbf{0.080} & 0.991 & \textbf{0.915} \\ \bottomrule
    \end{tabular}%
    }
    \caption{Accuracy ($\uparrow$), accuracy drop ($\uparrow$), and prediction-feature correlation ($\downarrow$) on four classification tasks of GPT-3 (Davinci, 175B), finetuned with and without explanations. The skewed training sets contain domain-specific cues.}
    \label{tab: domain}
\end{table*}

%% file: tables/corr_base_table.tex
\begin{table}[t]
\centering
\resizebox{\columnwidth}{!}{%
\begin{tabular}{lrrrr}
\toprule
 & \multicolumn{1}{l}{ComVE} & \multicolumn{1}{l}{CREAK} & \multicolumn{1}{l}{e-SNLI} & \multicolumn{1}{l}{SBIC} \\ \hline
Sentence Length & 0.018 & 0.056 & -0.114 & -0.226 \\
Present Tense & -0.022 & -0.010 & -0.004 & 0.135 \\
Embedding Cluster & 0.000 & -0.008 & 0.062 & 0.378 \\
Plural Noun & -0.062 & 0.006 & 0.007 & 0.112 \\
Dataset-specific & -0.051 & -0.004 & -0.059 & -0.068 \\ \bottomrule
\end{tabular}
}
\caption{Label-feature correlation in the unskewed training set $D_{train}$ without intentionally introduced spurious cues.}
\label{tab: corr base}
\end{table}

%% file: tables/standard_sample_outputs.tex
\begin{table*}[ht]
\centering
\resizebox{0.9\textwidth}{!}{%
\begin{tabularx}{\textwidth}{ccccXc}
\toprule
\begin{tabular}[c]{@{}l@{}}Plural\\ Filter\end{tabular} & \begin{tabular}[c]{@{}l@{}}Present \\ Tense\\ Filter\end{tabular} & \begin{tabular}[c]{@{}l@{}}Length\\ Filter\end{tabular} & \begin{tabular}[c]{@{}l@{}}Cluster\\ Filter\end{tabular} & Prompt & Completion \\ \hline
FALSE & FALSE & TRUE & FALSE & Is the following claim about Kidney true or false? Claim: The central organ for helping blood circulate is the Kidney. Answer: & false \\\vspace{0.2cm} 
TRUE & TRUE & FALSE & FALSE & Is the following claim about Eye true or false? Claim: Most humans use both Eyes together. Answer: & true \\\vspace{0.2cm} 
TRUE & FALSE & TRUE & TRUE & Is the following claim about Brothers Grimm true or false? Claim: Works by the Brothers Grimm were originally written in German. Answer: & true \\\vspace{0.2cm} 
FALSE & TRUE & FALSE & FALSE & Is the following claim about Street art true or false? Claim: Street art means graffiti which is illegal. Answer: & false \\ \vspace{0.2cm}
FALSE & FALSE & FALSE & TRUE & Is the following claim about Alan Greenspan true or false? Claim: Alan Greenspan was a comedian for a sketch show. Answer: & false \\ \bottomrule
\end{tabularx}
}
\caption{Davinci Outputs with Standard Finetuning}
\label{tab:my-table}
\end{table*}

%% file: tables/explain_sample_outputs.tex
\begin{table*}[ht]
\centering
\resizebox{0.9\textwidth}{!}{%
\begin{tabularx}{\textwidth}{ccccXX}
\toprule
\begin{tabular}[c]{@{}l@{}}Plural\\ Filter\end{tabular} & \begin{tabular}[c]{@{}l@{}}Present \\ Tense\\ Filter\end{tabular} & \begin{tabular}[c]{@{}l@{}}Length\\ Filter\end{tabular} & \begin{tabular}[c]{@{}l@{}}Cluster\\ Filter\end{tabular} & Prompt & Completion \\ \hline 
FALSE & FALSE & TRUE & FALSE & Is the following claim about Kidney true or false? Thoughts: & The organ that is central to circulating blood is the heart. Answer: false \\ \vspace{0.2cm}
TRUE & TRUE & FALSE & FALSE & Is the following claim about Eye true or false? Thoughts: & Both eyes work in tandem with the brain to process information about the world around them. Answer: true \\ \vspace{0.2cm}
TRUE & FALSE & TRUE & TRUE & Is the following claim about Brothers Grimm true or false? Thoughts: & This is true, they were German and from Hesse. Answer: true \\ \vspace{0.2cm}
FALSE & TRUE & FALSE & FALSE & Is the following claim about Street art true or false? Thoughts: & Street art is usually painted with permission or commissioned while graffiti is not. Answer: false \\ \vspace{0.2cm}
FALSE & FALSE & FALSE & TRUE & Is the following claim about Alan Greenspan true or false? Thoughts: & False, for the fact this is more of an outlandish statement and in no way true or factual in any basis. Answer: false \\
\bottomrule
\end{tabularx}
}
\caption{Davinci Outputs with Explanation Finetuning}
\label{tab:example-table}
\end{table*}